\documentclass[twocolumn]{mc_nankai}

\usepackage[utf8]{inputenc} 
\usepackage[T1]{fontenc}    
\usepackage{url}            
\usepackage{booktabs}       
\usepackage{amsfonts}       
\usepackage{nicefrac}       
\usepackage{microtype}      

\usepackage{graphicx}
\usepackage{wrapfig}
\usepackage{makecell}
\usepackage{amsmath}
\usepackage{float}
\usepackage{amssymb}
\usepackage{bm}
\usepackage{overpic}
\usepackage{multicol}
\usepackage{adjustbox}
\usepackage{multirow}
\usepackage{siunitx}
\usepackage{enumitem}
\usepackage{colortbl}
\usepackage{pifont}

\usepackage[dvipsnames,table,xcdraw]{xcolor}
\usepackage{caption}
\usepackage{subcaption}
\usepackage{tabularx}
\usepackage{arydshln}
\usepackage{algorithm}
\usepackage{algorithmic}
\usepackage{tikz}

\newcommand{\cmark}{\scalebox{1.5}{\checkmark}}
\newcommand{\xmark}{\scalebox{1.5}{\ding{55}}}

\def\pipeline{ASID-Verify}
\def\model{ASID-Captioner}  
\def\dataset{ASID-1M}

\title{Towards Universal Video MLLMs with Attribute-Structured and Quality-Verified Instructions}

\author[1]{Yunheng Li}
\author[1]{Hengrui Zhang}
\author[3]{Meng-Hao Guo}
\author[2]{Wenzhao Gao}
\author[2]{Shaoyong Jia}
\author[2]{\\Shaohui Jiao}

\author[1]{Qibin Hou$^{\dagger}$}
\author[1]{Ming-Ming Cheng}

\affiliation[1]{VCIP, School of Computer Science, Nankai University\\}
\affiliation[2]{ByteDance Inc.}
\affiliation[3]{Tsinghua University\\}

\affiliation[]{$^{\dagger}$Corresponding author.}

\project{https://asid-caption.github.io/}
\data{https://huggingface.co/datasets/AudioVisual-Caption/ASID-1M}
\mcmodel{https://huggingface.co/AudioVisual-Caption/models}
\sourcecode{https://github.com/HVision-NKU/ASID-Caption}
\date{\today}

\abstract{
Universal video understanding requires modeling fine-grained visual and audio information over time in diverse real-world scenarios.
However, the performance of existing models is primarily constrained by video-instruction data that represents complex audiovisual content as single, incomplete descriptions, lacking fine-grained organization and reliable annotation.
To address this, we introduce: (i) \dataset, an open-source collection of one million structured, fine-grained audiovisual instruction annotations with single- and multi-attribute supervision; 
(ii) \pipeline, a scalable data curation pipeline for annotation, with automatic verification and refinement that enforces semantic and temporal consistency between descriptions and the corresponding audiovisual content;
and (iii) \model, a video understanding model trained via Supervised Fine-Tuning (SFT) on the \dataset.
Experiments across seven benchmarks covering audiovisual captioning, attribute-wise captioning, caption-based QA, and caption-based temporal grounding show that \model~improves fine-grained caption quality while reducing hallucinations and improving instruction following.
It achieves state-of-the-art performance among open-source models and is competitive with Gemini-3-Pro.
}

\begin{document}

\maketitle
\justifying

\section{Introduction}
Multimodal large language models (MLLMs)~\cite{damonlpsg2023videollama,sun2024video,damonlpsg2024videollama2,zhang2025bee} play a central role in advancing universal video understanding by enabling unified modeling across visual, audio, and natural language modalities.
Nevertheless, understanding real-world videos remains challenging due to the temporal distribution and semantic diversity of fine-grained information.
In this context, the quality of video caption supervision plays a critical role, as expressive and well-structured captions not only improve captioning performance but also provide transferable semantic representations that support multimodal alignment and downstream video understanding and generation~\cite{hong2022cogvideo,chen2024sharegpt4video,zhang2025unified}.

\begin{figure}[t]
    \setlength{\abovecaptionskip}{2pt}
    \begin{center}
      \centerline{\includegraphics[width=0.98\columnwidth]{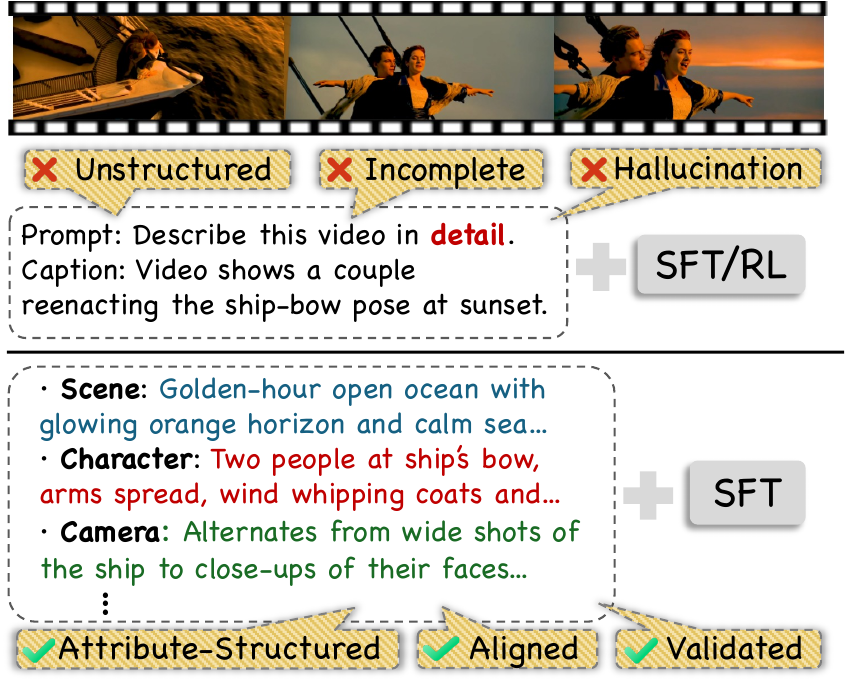}}
      \caption{Motivation. Existing caption supervision is typically unstructured, leading to incomplete descriptions. In contrast, our attribute-structured supervision aligns and validates each aspect against audiovisual evidence, enabling fine-grained learning.}
      \label{fig:intro}
    \end{center}
      \vskip -0.2in
  \end{figure}

Recent approaches~\cite{li2024llava,meng2025videocap,li2025videochat,li2025tempsamp,chen2025versavid,chen2025avocado} to video understanding typically rely on caption-based instruction data for supervised fine-tuning (SFT), often followed by reinforcement learning (RL) for behavior refinement, as illustrated in~\figref{fig:intro}.
Because SFT learns directly from captions, its effectiveness is determined by what information the captions explicitly provide.
While RL can adjust generation preferences, it cannot reliably reconstruct fine-grained information that is missing training supervision.
Thus, model performance is constrained by the availability and diversity of explicit, fine-grained supervision in the training data.

As summarized in~\tabref{tab:intro}, existing video instruction datasets provide limited support for learning fine-grained, controllable video understanding.
A common annotation paradigm assigns each video a single fixed-format prompt that produces an incomplete caption, often omitting critical fine-grained information such as temporal localization or camera-related details.
Importantly, our analysis in~\secref{sec: Analysis} demonstrates that increasing annotation granularity alone does not lead to higher-quality supervision: making multi-attribute captions more comprehensive through detailed prompt design or multi-source ensembling introduces additional incorrect and hallucinated content.
Such unreliable fine-grained annotations are unsuitable as training supervision.
However, prior datasets generally lack systematic verification for fine-grained annotations, primarily due to the high annotation cost and the difficulty of validating fine-grained annotations at scale.
Moreover, many recent fine-grained datasets remain closed, limiting reproducibility and preventing further verification and refinement of their annotations.

To address these challenges, we adopt a data-centric approach that focuses on structured instruction design and quality-verified fine-grained supervision for universal video understanding.
Specifically, we construct an open-source, million-scale audiovisual instruction dataset, in which annotations are decomposed into complementary single- and multi-attribute descriptions.
Rather than modeling each video as a single caption, our dataset adopts attribute-structured instructions for more explicit and fine-grained learning.
Moreover, we recognize that scaling fine-grained annotations requires principled control over their reliability and consistency.
To this end, we develop a multi-stage data curation pipeline that performs automatic annotation, verification, and refinement, enforcing semantic and temporal alignment with the corresponding audiovisual content.
Building on \dataset, we train \model~via a multi-stage supervised fine-tuning scheme, enabling improved fine-grained perception and instruction following.

We evaluate \model~on seven benchmarks that jointly assess caption quality, attribute-level instruction following, and the utility of captions for downstream reasoning and temporal localization.
Across these evaluations, \model~exhibits consistent improvements in fine-grained semantic coverage and instruction following over strong open-source models.
Notably, the 7B variant remains competitive with Gemini-3-Pro on several evaluations.
Our contributions are summarized as follows:
\begin{itemize}
  \item \dataset: An open-source, million-scale dataset of composable attribute-structured audiovisual instructions, providing fine-grained and complementary supervision for universal video understanding.
  \item \pipeline: A multi-stage data curation pipeline for annotation, automatic verification, and refinement, enforcing semantic and temporal consistency to improve the reliability of fine-grained supervision.
  \item \model: A video understanding model trained on \dataset, achieving consistent gains in fine-grained semantic coverage and attribute-level instruction following over strong open-source models.
\end{itemize}

\begin{table}[t]
    \centering
    \setlength{\abovecaptionskip}{2pt}
    \caption{Comparison of video instruction datasets in terms of annotation properties. $\dagger$ denotes datasets with both visual and auditory annotations, whereas others are visual-only.}
    \setlength{\tabcolsep}{3pt}
    \resizebox{\linewidth}{!}{
    \begin{tabular}{lccccc c}
    \toprule
    \textbf{Dataset} & \textbf{Fine-Grain} & \textbf{Multi-Attr} & \textbf{Verified} & \textbf{Open}\\
    \midrule
    Ego4D Narrations~\cite{grauman2022ego4d} & \xmark & \xmark & \xmark & \cmark \\
    InternVid~\cite{wang2023internvid} & \xmark & \xmark & \xmark & \cmark \\
    VideoInstruct-100K~\cite{Maaz2023VideoChatGPT} & \cmark & \xmark & \xmark & \cmark \\
    LLaVA-Video-178K~\cite{zhang2024video} & \cmark & \xmark & \xmark & \cmark \\
    ShareGPT4Video~\cite{chen2024sharegpt4video} & \cmark & \xmark & \xmark & \cmark \\
    FineVideo~\cite{Farr2024FineVideo} & \cmark & \xmark & \xmark & \cmark \\
    AVoCaDO$\dagger$~\cite{chen2025avocado} & \cmark & \xmark & \xmark & \xmark  \\
    OMNI-CAPTIONER$\dagger$~\cite{ma2025omni} & \cmark & \xmark & \xmark & \xmark  \\
    \midrule
    \dataset~$\dagger$ (Ours) & \cmark & \cmark & \cmark & \cmark \\
    \bottomrule
    \end{tabular}
    }
    \label{tab:intro}
    \vskip -0.1in
  \end{table}

\begin{figure*}[!t]
  \setlength{\abovecaptionskip}{2pt}
  \begin{center}
    \centerline{\includegraphics[width=0.95\linewidth]{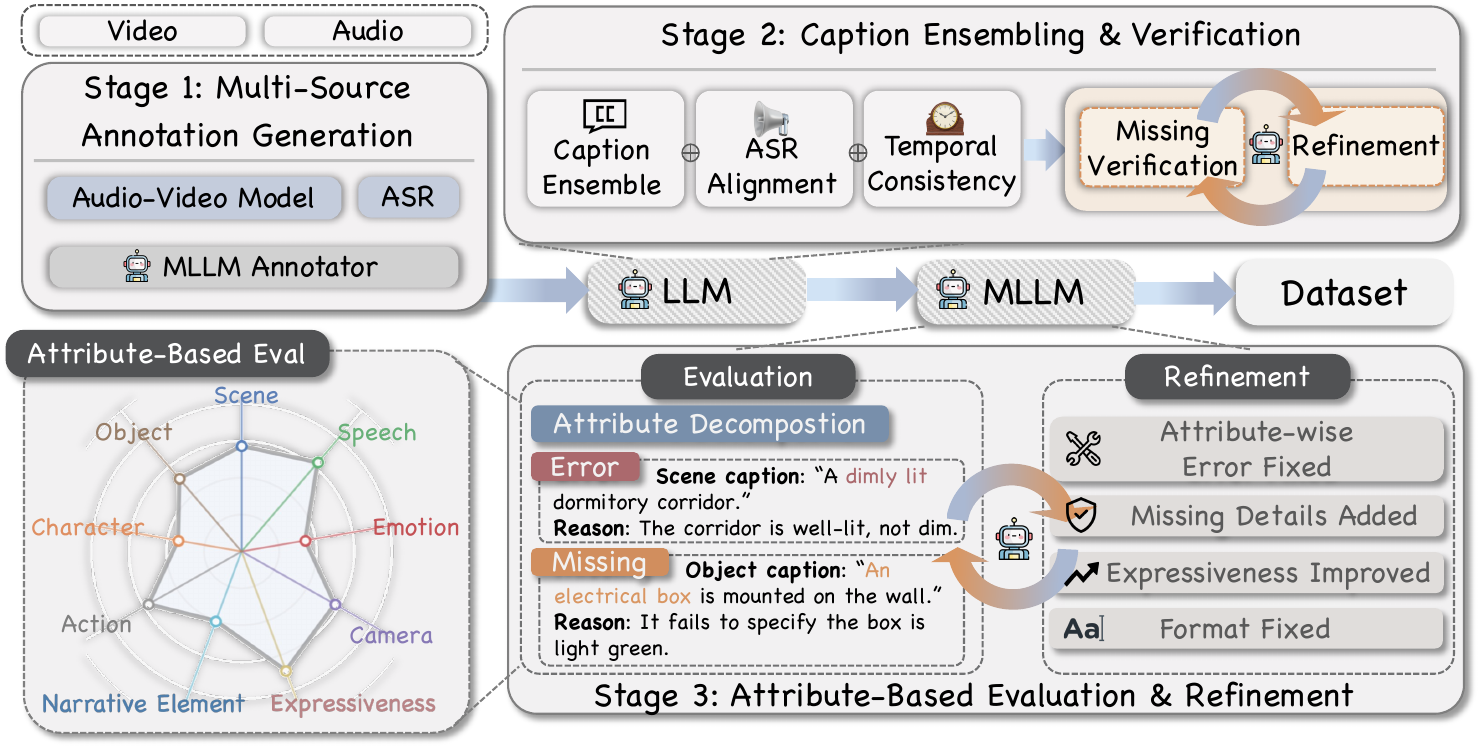}}
    \caption{Overview of \pipeline. Multi-source audiovisual annotations are first generated and ensembled with ASR alignment and temporal consistency verification. 
    Captions are then evaluated at the attribute level to identify missing or incorrect content and refined in a targeted manner, producing attribute-structured and quality-verified audiovisual instructions.}    
    \label{fig:pipeline}
  \end{center}
  \vskip -0.15in
\end{figure*}

\section{Related Work}

\subsection{Audiovisual Multimodal Models}
Recent advances in audiovisual multimodal models have extended large language models to video understanding through joint modeling of visual, audio, and language modalities.
Representative approaches jointly model visual representations, acoustic signals, and natural language, enabling instruction following, open-ended reasoning, and conversational interaction over video content~\cite{sun2024video,tang2025video,damonlpsg2023videollama,damonlpsg2024videollama2,Maaz2023VideoChatGPT,lin2024video,wang2022internvideo,song2024moviechat,wu2025ugc}.
Most existing methods are trained via supervised fine-tuning on instruction-style video-language data, learning to generate single monolithic responses conditioned on video inputs and textual prompts~\cite{wang2024tarsier,yuan2025tarsier2,chai2024auroracap}.
Beyond supervised learning, several works explore reinforcement learning or preference-based optimization to refine response behavior~\cite{li2025videochat,chen2025versavid,chen2025avocado,li2025tempsamp}.
However, these approaches operate mainly at the output level, leaving the semantic structure of supervision unchanged while often requiring carefully designed rewards and substantial training resources.
Thus, their effectiveness is constrained by the quality, granularity, and organization of instruction data, highlighting the central role of supervision design and motivating a systematic analysis of existing video instruction datasets.

\subsection{Datasets for Audiovisual Understanding}
Large-scale datasets have played a central role in advancing audiovisual understanding by supervised learning over diverse and complex video content.
Early efforts primarily focus on collecting video-text pairs, typically providing coarse video-level descriptions that support broad multimodal alignment~\cite{xu2016msr,zhou2018towards,krishna2017dense,bain2021frozen,wang2023internvid,miech2019howto100m}.
More recent instruction-style datasets extend this paradigm by pairing videos with conversational captions, increasing descriptive coverage and interaction flexibility, yet still providing limited annotation over fine-grained temporal structure and semantic composition~\cite{Maaz2023VideoChatGPT,zhang2024video,chen2024sharegpt4video,li2025if}.
To address these limitation, subsequent datasets explore finer-grained supervision by annotating specific aspects of video content, including actions, objects, and speech~\cite{grauman2022ego4d,Farr2024FineVideo,chen2025avocado}.
However, these fine-grained annotations are typically provided without a unified semantic structure, resulting in incomplete and inconsistent supervision across different aspects.
Moreover, the lack of systematic verification allows incorrect or hallucinated annotations to remain, limiting the effectiveness of fine-grained supervision.

\begin{figure*}[t]
    \centering
      \vspace{2pt}
    \setlength{\abovecaptionskip}{2pt}
    \includegraphics[width=1.0\linewidth]{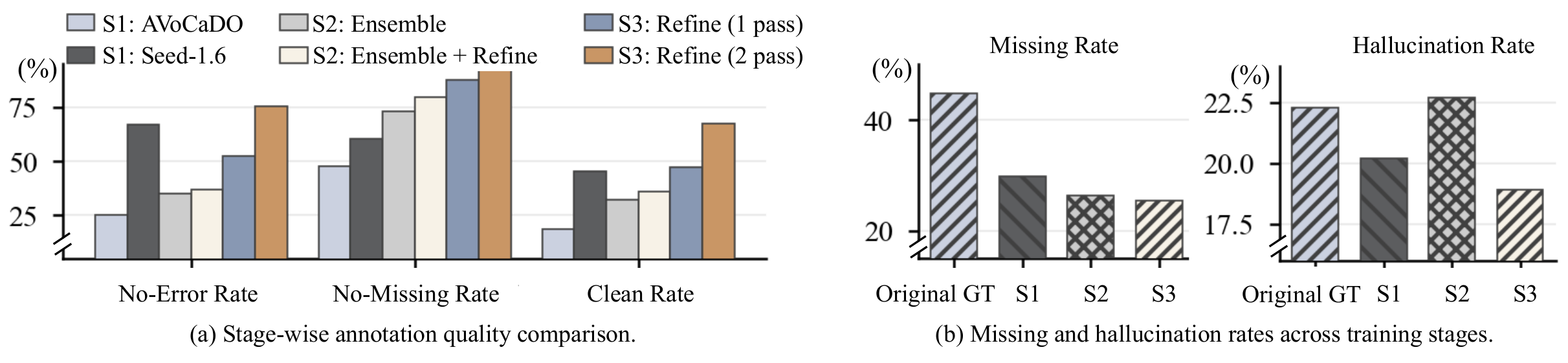}
    \caption{Stage-wise analysis of annotation quality and errors under progressively refined training data.}
    \label{fig:analysis}
\end{figure*}

\section{\dataset~and \pipeline~Pipeline}
To construct \dataset, we develop \pipeline, a scalable and reproducible data curation pipeline designed to address the limitations of existing audiovisual instruction datasets, particularly the lack of structured fine-grained annotation and systematic verification.
As shown in~\figref{fig:pipeline}, the pipeline adopts a staged process for instruction annotation that integrates annotation ensembling, validation, and targeted refinement instead of a single-pass scheme.
Each stage of this pipeline is detailed below.

\subsection{S1: Multi-Source Annotation Generation}
The pipeline begins by constructing an initial pool of approximately 125K videos from the YouTube subset of LLaVA-Video-178K~\cite{zhang2024video} and short-form videos (under 3 minutes) from FineVideo~\cite{Farr2024FineVideo} to ensure broad audiovisual coverage.
Initial annotations are generated using open-source audiovisual models (e.g., AVoCaDO~\cite{chen2025avocado}), which jointly model visual content and audio inputs.
Spoken content is transcribed and aligned via automatic speech recognition (ASR), using Whisper large-v3~\cite{radford2022whisper} for transcription and WhisperX~\cite{bain2022whisperx} for speaker-aware segmentation and utterance-level alignment.
The resulting ASR subtitles, together with explicit video playback timestamps, are provided as additional inputs to closed-source multimodal models (e.g., Seed-1.6\protect\footnotemark) that serve as additional annotators.
Notably, this design enables the multimodal models to produce fine-grained, multi-attribute captions that describe diverse aspects of the video content, while temporal grounding ensures precise timestamp-level alignment of these attributes.
Thus, for each video $v$, we collect multiple candidate captions $\{y^{(1)}_m(v)\}_{m=1}^{M}$ from these complementary annotators, where $M$ is the number of annotators.

\footnotetext{\url{https://seed.bytedance.com/en/seed1_6}}

\subsection{S2: Caption Ensembling and Verification}
Building on the multiple captions generated in Stage~1 from complementary annotation sources, this stage focuses on caption ensembling to produce more complete and fine-grained annotations.
Specifically, we employ a strong LLM (Seed-1.6) as an integrator to synthesize them into a unified all-attribute draft caption, denoted as $\tilde{y}^{(2)}_{\mathcal{A}}(v)$.
This integration prioritizes complementary coverage across sources, retaining fine-grained attribute details that may only appear in individual candidates, while resolving obvious redundancy or conflicts.
However, this initial integration may still introduce missing details or incorrect combinations.
To address this, the draft caption $\tilde{y}^{(2)}_{\mathcal{A}}(v)$ is further verified against the original audiovisual evidence.
In particular, we focus on ASR alignment to ground speech-related descriptions in speaker-aware transcripts, together with timestamp-level consistency checks across different semantic attributes.
Based on the verification results, selective refinement produces the caption $y^{(2)}_{\mathcal{A}}(v)$ by preserving validated content and removing inconsistent audiovisual descriptions, resulting in annotations that are both more comprehensive than individual source candidates.

\subsection{S3: Attribute-Based Evaluation and Refinement}
Although Stage~2 substantially improves captions with more complete and fine-grained attribute details, errors from source annotations and fusion can still accumulate.
A key challenge is that these errors are often entangled across multiple semantic attributes, which limits the effectiveness of caption-level verification.
Starting from the caption $y^{(2)}_{\mathcal{A}}(v)$, we decompose it into predefined semantic attributes and independently assess each attribute using an MLLM (Seed-1.6), as shown in~\figref{fig:pipeline}.
Beyond factual accuracy, we perform expressiveness-level validation to assess caption quality, including fluency, redundancy, and ambiguous references.
Attribute-wise verification is conducted under explicit rules for both {Error} and {Missing}, resulting in more precise and controllable assessment results.
Based on the evaluation results, refinement is selectively applied to the affected attributes while preserving the remaining content, thereby limiting error propagation and improving fine-grained annotation quality.
To ensure formatting consistency, auxiliary format validation and correction are applied to handle repetition collapse, malformed timestamps, and non-compliant symbols.
After attribute-based evaluation and refinement, 121K videos are retained as the final version of \dataset, each paired with eight verified single-attribute captions $\{y_a(v)\}_{a\in\mathcal{A}}$ and a final all-attribute caption $y_{\mathcal{A}}(v)$, which directly instantiate the progressive training supervision in Sec.~\ref{sec:training}.
In addition, a random subset of 300 videos is manually inspected: over 98\% of the refined captions are reliable, with remaining issues limited to minor timestamp deviations within 2 seconds.

\subsection{Analysis of Stage-Wise Contributions}
\label{sec: Analysis}
As shown in~\figref{fig:analysis}, we analyze stage-wise contributions using two evaluations: 
annotation quality measured by Stage~3 No-Error/No-Missing/Clean rates (left), and downstream effectiveness measured by training Qwen2.5-Omni-3B~\cite{xu2025qwen2} on 200K video captions and testing on Video-SALMONN-2~\cite{tang2025video} (right).
The results show that captions generated by a single model miss a substantial amount of fine-grained information, even for strong models such as AVoCaDO~\cite{chen2025avocado}. 
Models specialized in different aspects, such as visual description and speech recognition, tend to capture complementary information from the same video, and integrating their captions substantially improves semantic coverage.
Consistent with this observation, early stages that aggregate captions from multiple sources improves coverage, which is reflected by higher No-Missing rates and lower Missing rates after ensembling.
However, we find that caption ensembling itself is inherently challenging: single-pass fusion often fails to preserve all fine-grained details present in the source captions.
As finer-grained descriptions are incorporated, the ensembling process increasingly introduces Hallucinated content and exposes annotation errors inherited from earlier data sources that were not systematically addressed by prior datasets.
Stage~3 provides the key improvement by introducing attribute-level evaluation and targeted refinement, which further reduces missing, effectively suppresses hallucination, and improves the Clean rate.

\begin{figure}
    \centering
    \setlength{\abovecaptionskip}{2pt}
    \includegraphics[width=1\linewidth]{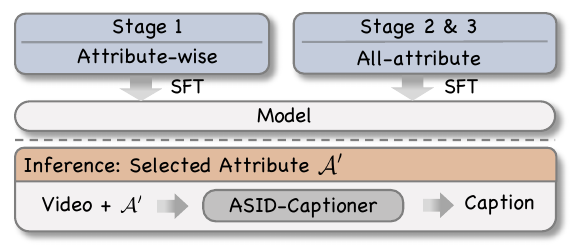}
    \caption{Overview of progressive attribute learning with stage-wise training and controllable attribute selection at inference.}
    \label{fig:training_recipe}
\end{figure}

\section{Progressive Attribute Learning}
\label{sec:training}
As shown in~\figref{fig:training_recipe}, we adopt a three-stage training scheme that progressively expands from single-attribute to all-attribute supervision, enabling stable optimization and improved generalization at inference.
Our model is built upon Qwen2.5-Omni~\cite{xu2025qwen2} and optimized via SFT.

\begin{table*}[t]
    \centering
    \small
    \setlength{\abovecaptionskip}{2pt}
    \setlength{\tabcolsep}{4pt}
    \caption{Model performance on the audiovisual captioning benchmarks. Evaluation is conducted using GPT-4.1 as the judge model for Video-SALMONN-2 and GPT-4o for UGC-VideoCap. $^*$Denotes closely related work.}
    \resizebox{\linewidth}{!}{
    \begin{tabular}{lccccccccc}
    \toprule
        \multirow{2}{*}[-0.1cm]{\textbf{Model}} & \multirow{2}{*}[-0.1cm]{\textbf{Size}} & \multirow{2}{*}[-0.1cm]{\textbf{Modality}} & \multicolumn{3}{c}{\textbf{video-SALMONN-2 testset}} & \multicolumn{4}{c}{\textbf{UGC-VideoCap}} \\
        \cmidrule(lr){4-6} \cmidrule(lr){7-10}
        ~ & ~ & ~ & Miss~$\downarrow$ & Hall.~$\downarrow$ & Total~$\downarrow$ & Audio~$\uparrow$ & Visual~$\uparrow$ & Detail~$\uparrow$ & Avg.~$\uparrow$ \\
        \midrule
        \textcolor{gray!100}{Gemini-3-Pro} & - & A + V & 19.1 & 16.6 & 35.7 & 80.4 & 84.7 & 80.6 & 81.9 \\
        \textcolor{gray!100}{Gemini-2.5-Pro} & - & A + V & 18.1 & 13.3 & 31.3 & 69.5 & 74.7 & 73.7& 72.6 \\
        \textcolor{gray!100}{Gemini-2.5-Flash} & - & A + V & 19.3 & 13.9 & 33.3 & 69.1 & 75.8 & 74.0 & 73.0 \\
        InternVL3.5~\cite{wang2025internvl3} & 8B & V & 53.8 & 25.5 & 79.4 & 47.9 & 64.8 & 59.5 & 57.4 \\
        Qwen2.5-VL~\cite{bai2025qwen2} & 7B & V & 40.5 & 17.0 & 57.5 & 46.6 & 69.1 & 62.3 & 59.3 \\
        HumanOmniV2~\cite{yang2025humanomniv2} & 7B & A + V & 49.2 & {\textbf{12.3}} & 61.6 & 45.6 & 66.3 & 59.5 & 57.1 \\
        ARC-Hunyuan-Video~\cite{ge2025arc} & 7B & A + V & 45.7 & \underline{12.5} & 58.2 & 52.7 & 56.0 & 55.8 & 54.8 \\
        Qwen2.5-Omni~\cite{xu2025qwen2} & 7B & A + V & 41.7 & 15.4 & 57.1 & 46.9 & 66.1 & 60.0 & 57.7 \\
        MiniCPM-o-2.6~\cite{MiniCPM-o} & 8B & A + V & 42.2 & 14.3 & 56.5 & 38.6 & 68.5 & 57.7 & 54.9 \\
        video-SALMONN-2$^*$~\cite{tang2025video} & 7B & A + V & 21.2 & 17.6 & {38.8} & {61.8} & {71.4} & {68.5} & {67.2} \\
        UGC-VideoCaptioner$^*$~\cite{wu2025ugc} & 3B & A + V & 31.6 & 17.0 & 48.6 & 61.4 & 58.4 & 57.5 & 59.1 \\
        {Qwen3-Omni-Instruct~\cite{xu2025qwen3}} & {30B-A3B} & {A + V} & {32.0} & {13.6} & {45.6} & {67.5} & {74.8} & \underline{72.3} & {71.5} \\
        {Qwen3-Omni-Captioner~\cite{xu2025qwen3}} & {30B-A3B} & {A + V} & {31.0} & {16.6} & {47.6} & {69.0} & {75.5} & \underline{72.3} & {72.3} \\
        AVoCaDO$^*$~\cite{chen2025avocado}  & 7B & A + V & \underline{21.1} & 16.2 & \underline{37.3} & {73.0} & {74.6} & {71.8} & \underline{73.2} \\
        \rowcolor{gray!12} \model~(Ours) & 3B & A + V & {23.4} & {18.3} & {41.7} & \underline{78.6} & \textbf{84.8} & \textbf{80.2} & \textbf{81.2} \\
        \rowcolor{gray!12} \model~(Ours) & 7B & A + V & \textbf{20.5} & {15.4} & \textbf{35.9} & \textbf{79.1} & \underline{84.4} & \textbf{80.2} & \textbf{81.2} \\
    \bottomrule
    \end{tabular}
    }
    \addtocounter{footnote}{-1}
    \label{tab:avcap result}
  \end{table*}

  \begin{table*}[!t]
      \setlength{\abovecaptionskip}{2pt}
      \caption{Comparison of attribute-wise visual captioning performance on the VDC benchmark.
      Accuracy and judge scores are reported for each visual aspect, with Seed-1.6 used as the judge model.}  
      \centering
      \setlength{\tabcolsep}{1pt}
      \resizebox{1.0\textwidth}{!}{\begin{tabular}{l ccccccc }
      \toprule
      
      \multirow{2}{*}{Model} & \multirow{2}{*}{Size} &\multicolumn{1}{c}{Camera} &
      \multicolumn{1}{c}{Short}  & \multicolumn{1}{c}{Background} & \multicolumn{1}{c}{Main Object} & \multicolumn{1}{c}{Detailed}  &\multicolumn{1}{c}{Avg.} \\
      \cmidrule(lr){3-3}
      \cmidrule(lr){4-4} 
      \cmidrule(lr){5-5} 
      \cmidrule(lr){6-6} 
      \cmidrule(lr){7-7} 
      \cmidrule(lr){8-8} 
      
      && ~Acc~/~Score~$\uparrow$ & ~Acc~/~Score~$\uparrow$ & ~Acc~/~Score~$\uparrow$  & ~Acc~/~Score~$\uparrow$ & ~Acc~/~Score~$\uparrow$& ~Acc~/~Score~$\uparrow$ \\
      \midrule
      \textcolor{gray!100}{Gemini-3-Pro} & -& 35.5~/~1.5~& 27.6~/~1.2& 42.3~/~1.9 & {45.4~/~2.0}& 40.2~/~1.8~& 38.2~/~1.7 \\
      \textcolor{gray!100}{Gemini-2.5-Pro} & - & 35.0~/~1.5& 26.6~/~1.1& 41.5~/~1.8& 42.3~/~1.8& 38.5~/~1.7& 36.8~/~1.6 \\
        \textcolor{gray!100}{Gemini-2.5-Flash} & - & 34.5~/~1.4& 26.5~/~1.1& 41.1~/~1.7& 41.6~/~1.8& 38.1~/~1.7& 36.4~/~1.5 \\
        Qwen2.5-Omni~\cite{xu2025qwen2} & 3B &   28.0~/~1.3& 23.2~/~1.2& 35.1~/~1.6& 34.6~/~1.7&  32.4~/~1.4& 30.7~/~1.4 \\
        Qwen2.5-Omni~\cite{xu2025qwen2} & 7B &   29.5~/~1.3& 26.1~/~1.2& 38.2~/~1.7& 37.3~/~1.7& 34.2~/~1.5& 33.1~/~1.5 \\
      ARC-Hunyuan-Video~\cite{ge2025arc} & 7B & 29.1~/~1.3& 25.5~/~1.2& 37.4~/~1.6& 36.5~/~1.5& 34.3~/~1.5& 32.6~/~1.4  \\
      Qwen3-Omni-Instruct~\cite{xu2025qwen3} & {30B-A3B} & 33.7~/~1.4& 26.3~/~1.1& 41.8~/~1.9& 42.7~/~1.9&38.2~/~1.6& 36.5~/~1.6  \\
      Qwen3-Omni-Captioner~\cite{xu2025qwen3} & {30B-A3B} & 34.4~/~1.4& 26.8~/~1.2& 42.1~/~1.9& \underline{44.7~/~2.0}& 39.4~/~1.7& 37.5~/~1.6   \\
      AvoCaDO~\cite{chen2025avocado} & 7B& 35.3~/~1.5 & 27.1~/~1.2 & 41.0~/~1.8 & {40.0}~/~{1.8} & 37.9~/~1.7 & 36.3~/~1.6\\
      \rowcolor{gray!12} \model~(Ours) & 3B& \underline{37.0~/~1.6}& \underline{28.4~/~1.3}& \underline{45.5~/~2.0}& \underline{44.7~/~2.0}& \underline{41.7~/~1.8}& \underline{39.5~/~1.7}  \\
      \rowcolor{gray!12} \model~(Ours) & 7B & \textbf{38.2~/~1.7}& \textbf{28.8~/~1.3}& \textbf{46.9~/~2.1}& \textbf{47.4~/~2.1}& \textbf{43.2~/~1.9}& \textbf{40.9~/~1.8} \\
      \bottomrule
      \end{tabular}}
      \label{tab:structured_baseline}
  \vskip -0.1in
      \end{table*}

\myPara{Stage 1: Attribute-Wise Representation Learning.}
Training is conducted under single-attribute supervision with attribute-conditioned targets.
Each training instance comprises a video $v$, an attribute $a \in \mathcal{A}$, where $\mathcal{A}$ denotes the predefined attribute set, and a corresponding attribute-specific target caption $y_a$ describing the aspect specified by $a$.
The model is optimized by minimizing the negative log-likelihood:
\begin{equation}
\mathcal{L}_1
=
\mathbb{E}_{(v,a,y_a)\sim\mathcal{D}_1}
\left[
-\log p(y_a \mid v, a)
\right],
\end{equation}
where $\mathcal{D}_1$ denotes the collection of training samples $(v,a,y_a)$ used in Stage~1.
This stage restricts supervision to a single attribute per instance, simplifying the learning objective and enforcing attribute-specific learning.

\myPara{Stage 2: All-Attribute Learning with Short Context.}
Stage~2 replaces single-attribute supervision with all-attribute supervision on short video clips, requiring the model to jointly model multiple attributes.
Each training instance is associated with the full attribute set $\mathcal{A}$ and a joint target caption $y_{\mathcal{A}}$.
The model is optimized by minimizing:
\begin{equation}
  \mathcal{L}_2
  =
  \mathbb{E}_{(v,y_{\mathcal{A}})\sim\mathcal{D}_2}
  \left[
  -\log p(y_{\mathcal{A}} \mid v, \mathcal{A})
  \right].
\end{equation}  
where $\mathcal{D}_2$ denotes the Stage~2 training samples supervised with all-attribute captions.
This stage shifts learning from attribute-wise modeling to joint modeling over all attributes under a short temporal context.

\myPara{Stage 3: Long-Context All-Attribute Learning.}
Stage~3 trains the model on long video clips (up to 3 minutes) under all-attribute supervision.
Compared to Stage~2, the learning objective remains unchanged, while training is conducted on a distinct data distribution $\mathcal{D}_3$ that exposes the model to extended temporal context.

\myPara{Inference.}
At inference time, the model accepts a video $v$ together with a user-specified set of attributes $\mathcal{A}' \subseteq \mathcal{A}$ and generates a caption conditioned on the selected attributes.
This enables flexible control over caption content by choosing different attribute combinations as input, without requiring retraining or architectural changes.

\section{Experiments}

\subsection{Benchmarks}

%
We evaluate \model~on seven benchmarks that offer complementary perspectives on caption quality.
\textbf{Audiovisual captioning:} We report results on video-SALMONN-2 testset~\cite{tang2025video} and UGC-VideoCap~\cite{wu2025ugc} (\tabref{tab:avcap result}). 
video-SALMONN-2 explicitly measures caption reliability by penalizing missing (Miss) and hallucinations (Hall.), while UGC-VideoCap assesses modality-aware caption quality with separate scores for audio, visual, and detail.
\textbf{Attribute-wise visual captioning:} To evaluate fine-grained visual caption quality, we adopt an aspect-wise protocol on the VDC benchmark~\cite{chai2024auroracap} (\tabref{tab:structured_baseline}) with five fields, including camera, short summary, background, main object, and detailed description, and report both accuracy and judge scores.
\textbf{Text-to-video generation caption:} We evaluate on VidCapBench-AE~\cite{chen2025vidcapbench} (\tabref{tab:VidCapBench results}), which measures caption quality using QA pairs aligned with text-to-video generation across four aspects: video aesthetics, video content, video motion, and physical laws, and reports accuracy (Acc), precision (Pre), coverage (Cov), and conciseness (Con) for each aspect.
\textbf{Caption-based QA:} We assess whether captions preserve sufficient evidence for reasoning on Daily-Omni~\cite{zhou2025daily} and World-Sense~\cite{hong2025worldsense} using a caption-to-QA protocol (\tabref{tab:qa result}).
\textbf{Caption-based temporal grounding:}
We evaluate whether captions support temporal localization on Charades-STA~\cite{gao2017tall} (\tabref{tab:grounding result}).
Start and end timestamps of the queried moment are predicted using only the caption, and we report mIoU and recall at IoU thresholds (R1@$\tau$).

\begin{table*}[t]
  \centering
  \setlength{\abovecaptionskip}{2pt}
  \caption{Results on the VidCapBench-AE benchmark.
  Acc, Pre, Cov, and Con denote accuracy, precision, coverage, and conciseness, respectively, with Con scaled by $100$.
  Scores are computed using GPT-4o following \cite{chai2024auroracap}.}
  \setlength{\tabcolsep}{3pt}
  \resizebox{\linewidth}{!}{
  \begin{tabular}{lcccccc}
  \toprule
      \multirow{2}{*}{\textbf{Model}} & \multirow{2}{*}{\textbf{Size}}  & \textbf{Overall} & \textbf{Video Aesthetics} & \textbf{Video Content} & \textbf{Video Motion} & \textbf{Physical Laws} \\
      \cmidrule(lr){3-3}
      \cmidrule(lr){4-4} 
      \cmidrule(lr){5-5} 
      \cmidrule(lr){6-6} 
      \cmidrule(lr){7-7} 
      ~ & ~ & \textbf{Acc$\uparrow$/Pre$\uparrow$/Cov$\uparrow$/Con$\downarrow$} & \textbf{Acc$\uparrow$/Pre$\uparrow$/Cov$\uparrow$/Con$\downarrow$}& \textbf{Acc$\uparrow$/Pre$\uparrow$/Cov$\uparrow$/Con$\downarrow$} & \textbf{Acc$\uparrow$/Pre$\uparrow$/Cov$\uparrow$/Con$\downarrow$} &\textbf{Acc$\uparrow$/Pre$\uparrow$/Cov$\uparrow$/Con$\downarrow$} \\
      \toprule
      \textcolor{gray!100}{Gemini-3-Pro} & - & 18.9 / 58.1 / 92.8 / ~3.7~ & 16.7 / 51.4 / 89.8 / ~3.3~ & 18.9 / 61.5 / 94.1 / ~3.7~ & 10.7 / 39.8 / 91.6 / ~2.1~ & 38.2 / 58.3 / 93.5 / ~7.5~ \\
      \textcolor{gray!100}{Gemini-1.5-Pro-002} & - & 17.1 / 54.8 / 87.4 / ~9.2~ & 16.4 / 47.6 / 85.4 / ~8.8~ & 16.9 / 57.8 / 88.5 / ~9.1~ & ~9.8~ / 45.1 / 80.9 / ~5.3~ & 28.4 / 59.3 / 88.2 / 15.3 \\
      \textcolor{gray!100}{GPT-4o-20240806} & - & 16.8 / 57.4 / 86.0 / ~5.9~ & 14.1 / 47.6 / 83.4 / ~4.9~ & 17.5 / 61.7 / 87.2 / ~6.1~ & 10.2 / 41.3 / 84.0 / ~3.6~ & 27.9 / 62.1 / 85.4 / ~9.7~ \\
      Pixtral~\cite{agrawal2024pixtral} & 124B & 13.0  / 48.3 / 80.5 / ~3.0~ & 13.9 / 44.6 / 80.2 / ~3.2~ & 11.9 / 50.0 / 80.5 / ~2.7~ & ~6.2~ / 28.3 / 81.8 / ~1.4~ & 27.9 / 55.9 / 83.2 / ~6.4~ \\
      InternVL2~\cite{chen2024far} & 76B & ~7.4~  / 35.6 / 78.9 / \textbf{~0.7~} & ~5.8~ / 27.6 / 76.2 / \textbf{~0.6~} & ~7.2~ / 38.1 / 80.1 / \textbf{~0.7~} & ~4.4~ / 24.4 / 80.0 / \textbf{~0.4~} & 23.1 / 55.0 / 78.1 / \textbf{~2.3~} \\
      Qwen2-VL~\cite{wang2024qwen2} &72B & 12.2 / 46.8 / 79.0 / ~7.7~ & 12.0 / 42.5 / 79.2 / ~7.6~ & 11.5 / 48.4 / 78.8 / ~7.3~ & ~5.8~ / 28.6 / 77.8 / ~3.7~ & 27.1 / 59.6 / 80.9 / 17.2 \\
      Tarsier~\cite{wang2024tarsier} & 34B & 13.5 / 50.8 / 82.1 / 15.1 & 14.7 / 43.9 / 85.5 / 16.4 & 12.4 / 53.7 / 80.4 / 13.8 & ~7.1~ / \underline{38.1} / 84.0 / ~7.9~ & 28.1 / \textbf{61.7} / 83.9 / 31.4 \\
      Aria~\cite{li2024aria} & 25B & 14.1 / 51.5 / 84.4 / ~4.5~ & 13.0 / 44.0 / 82.7 / ~4.2~ & 13.9 / 54.9 / 85.3 / ~4.4~ & ~7.1~ / 34.2 / 81.8 / ~2.3~ & 27.9 / 56.8 / 83.7 / ~8.9~ \\
      Pixtral~\cite{agrawal2024pixtral} & 12B & 11.0 / 39.5 / 79.6 / ~5.2~ & 14.5 / 42.7 / 82.8 / ~6.8~ & ~8.6~ / 37.9 / 78.4 / ~4.0~ & ~3.6~ / 18.6 / 69.3 / ~1.7~ & 28.6 / 52.4 / 82.4 / 13.5 \\
      Llava-Next-Video~\cite{zhang2024llavanext-video} & 7B & 10.6 / 42.3 / 79.4 / ~3.9~ & 11.3 / 39.9 / 82.2 / ~4.2~ & ~9.6~ / 43.2 / 78.1 / ~3.5~ & ~4.4~ / 23.7 / 75.1 / ~1.7~ & 24.4 / 54.5 / 82.9 / ~9.0~ \\
      LongVA~\cite{zhang2024long} & 7B & 10.8 / 43.0 / 79.3 / ~6.1~ & 12.8 / 42.1 / 83.8 / ~7.3~ & ~9.2~ / 43.4 / 77.2 / ~5.2~ & ~4.9~ / 25.1 / 79.6 / ~2.8~ & 24.9 / 52.9 / 83.2 / 14.1 \\
      mPLUG-Owl3~\cite{ye2024mplug} & 7B & \underline{14.5} / 49.6 / 84.4 / ~6.9~ & 12.9 / 40.7 / 83.7 / ~6.1~ & \underline{14.8} / 53.5 / 85.1 / ~7.0~ & ~5.3~ / 33.3 / 80.0 / ~2.5~ & 26.9 / 55.7 / 81.7 / 12.8 \\
      InternVL2~\cite{chen2024far} & 8B & 10.2 / 43.0 / \underline{84.9} / \underline{~2.5~} & ~9.1~ / 36.3 / 84.4 / \underline{~2.2~} & 10.0 / 46.1 / 85.2 / \underline{~2.4~} & ~4.4~ / 18.0 / 81.3 / \underline{~1.1~} & 23.6 / 52.8 / \underline{85.7} / \underline{~5.8~} \\
      Qwen2-VL~\cite{wang2024qwen2} & 7B & 11.1  / 47.1 / 77.0 / ~6.4~ & 12.4 / 44.3 / 78.7 / ~7.2~ & ~9.9~ / 48.3 / 75.9 / ~5.7~ & ~4.0~ / 22.7 / 78.2 / ~2.3~ & 26.1 / 59.4 / 81.2 / 15.1 \\
      CogVLM2-Caption~\cite{yang2024cogvideox} & 5B & 13.1 / 49.2 / 85.1 / ~8.4~ & 12.5 / 45.2 / 83.1 / ~8.0~ & 12.7 / 50.8 / 86.3 / ~8.1~ & ~5.7~ / 33.9 / 82.7 / ~3.7~ & 27.9 / 59.9 / 82.7 / 17.8 \\
      \rowcolor{gray!12} \model~(Ours) & 3B& \textbf{18.2} / \underline{58.3} / \underline{93.0} / ~3.6~ & \textbf{15.5} / \textbf{50.7} / \underline{90.8} / ~3.1~ & \textbf{18.7} / \underline{62.5} / \underline{94.2} / ~3.7~ & \underline{8.9} / 34.0 / \textbf{90.6} / ~1.8~ & \textbf{33.7} / 55.6 / \underline{89.5} / ~6.7~ \\
      \rowcolor{gray!12} \model~(Ours) & 7B& \textbf{18.2} / \textbf{60.0} / \textbf{93.3} / ~3.7~ & \underline{15.4} / \underline{49.7} / \textbf{91.2} / ~3.1~ & \textbf{18.7} / \textbf{64.8} / \textbf{94.5} / ~3.8~ & \textbf{12.9} / \textbf{42.7} / \underline{88.6} / ~2.6~ & \underline{32.5} / \underline{60.2} / \textbf{91.7} / ~6.6~ \\
      \bottomrule
  \end{tabular}
  }
  \label{tab:VidCapBench results}
  \vskip -0.1in
\end{table*}



\begin{table}[t]
  \centering
  \setlength{\abovecaptionskip}{2pt}
  \caption{QA performance by Gemini-2.5-Pro based on captions.
  }
  \setlength{\tabcolsep}{1pt}
  \resizebox{\linewidth}{!}{
  \begin{tabular}{lccc}
  \toprule
      \textbf{Model} & \textbf{Size} & \textbf{{Daily-Omni$\uparrow$}} & \textbf{{World-Sense$\uparrow$}} \\
      \midrule
      \textcolor{gray!100}{Gemini-2.5-Pro} & - & 60.2 & 33.8 \\
      \textcolor{gray!100}{Gemini-2.5-Flash} & - & 55.3 & 31.0 \\
      HumanOmniV2~\cite{yang2025humanomniv2} & 7B & ~8.2~ & ~6.6~ \\
      ARC-Hunyuan-Video~\cite{ge2025arc} & 7B & ~8.6~ & ~8.7~ \\
      MiniCPM-o-2.6~\cite{MiniCPM-o} & 8B & ~9.8~ & ~7.2~ \\
      Qwen2.5-Omni~\cite{xu2025qwen2} & 7B & 13.4 & ~8.6~ \\
      UGC-VideoCaptioner~\cite{wu2025ugc} & 3B & 17.0 & 11.2 \\
      video-SALMONN-2~\cite{tang2025video} & 7B & 29.9 & 18.2 \\
      {Qwen3-Omni-Instruct}~\cite{xu2025qwen3} & {30B-A3B} & {17.5} & {12.7} \\
      {Qwen3-Omni-Captioner}~\cite{xu2025qwen3} & {30B-A3B} & {27.2} & {14.1} \\
      AVoCaDO~\cite{chen2025avocado}  & 7B & {50.1} & {25.7} \\
      \rowcolor{gray!12} \model~(Ours) & 3B & \underline{55.5} & \underline{32.3} \\
      \rowcolor{gray!12} \model~(Ours) & 7B & \textbf{61.2} & \textbf{34.0} \\
  \bottomrule
  \end{tabular}
  }
  \label{tab:qa result}
  \vskip -0.05in
\end{table}


\subsection{Main Results}

\myPara{Audiovisual captioning.}
We evaluate audiovisual captioning on Video-SALMONN-2 and UGC-VideoCap, which assess complementary aspects of caption quality.
As shown in~\tabref{tab:avcap result}, \model~achieves a strong balance between caption completeness and reliability across both benchmarks.
On Video-SALMONN-2, \model~achieves lower missing rates while maintaining controlled hallucination levels, resulting in improved caption reliability.
In contrast, several general-purpose models, such as InternVL3.5 and Qwen2.5-VL, produce conservative captions with fewer hallucinations but substantially more missing details.
On UGC-VideoCap, \model~performs consistently well across audio, visual, and detailed description dimensions, outperforming prior open-source omni models including UGC-VideoCaptioner and AVoCaDO.
Overall, \model~achieves performance comparable to recent Gemini models while maintaining a competitive balance of caption completeness and reliability among open-source captioning and omni models.

\begin{table}[t]
  \centering
  \setlength{\abovecaptionskip}{2pt}
  \caption{Temporal grounding by Gemini-2.5-Pro based on captions.
  }
  \resizebox{\linewidth}{!}{
  \begin{tabular}{lccccc}
  \toprule
      \textbf{Model}& \textbf{Size} & \textbf{mIoU$\uparrow$} & \textbf{{R1@0.3$\uparrow$}} & \textbf{{R1@0.5$\uparrow$}}  & \textbf{{R1@0.7$\uparrow$}}\\
      \midrule
      \textcolor{gray!100}{Gemini-3-Pro} & - & 8.8 & 46.4 & 26.0 & 10.2 \\
      \textcolor{gray!100}{Gemini-2.5-Pro} & - & 27.6 & 45.7 & 25.6 & 8.7 \\
      \textcolor{gray!100}{Gemini-2.5-Flash} & - & 6.1 & 40.0 & 19.5 & 7.1 \\
      \rowcolor{gray!12} \model~(Ours) & 3B & {24.1} & {38.0}  & {20.0}  & {8.0} \\
      \rowcolor{gray!12} \model~(Ours) & 7B & \textbf{28.5} & \textbf{44.1}& \textbf{26.3} & \textbf{11.0}  \\
  \bottomrule
  \end{tabular}
  }
  \label{tab:grounding result}
  \vskip -0.1in
\end{table}

\myPara{Attribute-wise visual captioning.}
\tabref{tab:structured_baseline} reports attribute-wise results under the five-field protocol.
\model~performs consistently well across all attributes, with clear advantages on camera and main object, where fine-grained visual details are often underrepresented by previous models.
In contrast, large-scale MoE-based models improve overall coverage but do not consistently translate their capacity into stronger attribute-wise accuracy.
We further observe that \model~is competitive with Gemini models on this benchmark and surpasses AVoCaDO in averaged attribute-wise performance, demonstrating robust fine-grained visual captioning.

\myPara{Text-to-video generation caption.}
We evaluate caption quality on VidCapBench-AE, which assesses how well captions support text-to-video generation across video aesthetics, content, motion, and physical laws.
\tabref{tab:VidCapBench results} shows that \model~achieves strong overall performance across all aspects, with consistently higher precision and coverage compared to most existing approaches.
Notably, the gains are most pronounced for video content and video motion, indicating more accurate descriptions of events and dynamics that are critical for text-to-video generation.
Moreover, \model~remains competitive on video aesthetics and physical laws, while performance on aesthetics and physical laws remains competitive.

\myPara{Caption-grounded QA.}
\tabref{tab:qa result} evaluates whether captions preserve sufficient evidence for downstream reasoning by constraining a fixed QA model (Gemini-2.5-Pro) to answer questions using captions only.
Under this setting, \model~achieves the strongest performance on both Daily-Omni and World-Sense, outperforming prior open-source captioning and omni models by a clear margin.
The advantage is more pronounced on World-Sense, which requires reasoning over entities, interactions, and speech, rather than relying on isolated visual evidence.
Both Daily-Omni and World-Sense are fully multimodal benchmarks, and \model~maintains strong performance on both the 3B and 7B variants, indicating that the gains stem from higher-quality audiovisual captions rather than model scale.

\myPara{Caption-based temporal grounding.}
Similar to caption-grounded QA, \tabref{tab:grounding result} evaluates temporal grounding by constraining a strong pretrained model (Gemini-2.5-Pro) to predict the start and end time of the queried event using captions only.
Under this setting, \model~achieves strong performance across all metrics, with clear improvements at higher IoU thresholds.
This indicates that the generated captions preserve temporally precise information rather than coarse event references.

\begin{table}[t]
  \centering
  \small
  \setlength{\abovecaptionskip}{2pt}
  \setlength{\tabcolsep}{12pt}
  \caption{Ablation of supervision data variants for captioning.}
  \resizebox{\linewidth}{!}{
  \begin{tabular}{lccccccccc}
  \toprule
      \multirow{2}{*}[-0.1cm]{\textbf{Dataset}} & \multicolumn{3}{c}{\textbf{video-SALMONN-2 testset}} & \multicolumn{1}{c}{\textbf{DVC Detailed}} \\
      \cmidrule(lr){2-4}
      \cmidrule(lr){5-5}
      ~ & Miss~$\downarrow$ & Hall.~$\downarrow$ & Total~$\downarrow$ &  ~Acc~/~Score~~$\uparrow$ \\
      \midrule
      - & 44.8 & 16.4& 61.2& 31.9 / 1.4\\
      Original GT & 43.9&22.3&66.2 & 35.5 / 1.5\\
      Non-attribute & 30.6& 18.5&49.1&38.5 / 1.6\\
      Multi-attribute & \textbf{25.5} & \textbf{18.9}& \textbf{43.4}& \textbf{40.2} / \textbf{1.7}\\
  \bottomrule
  \end{tabular}
  }
  \addtocounter{footnote}{-1}
  \label{tab:abla dataset}
  \vskip -0.1in
\end{table}


\begin{figure*}[t]
  \centering
  \vskip 0.05in
  \setlength{\abovecaptionskip}{2pt}
  \includegraphics[width=1.0\linewidth]{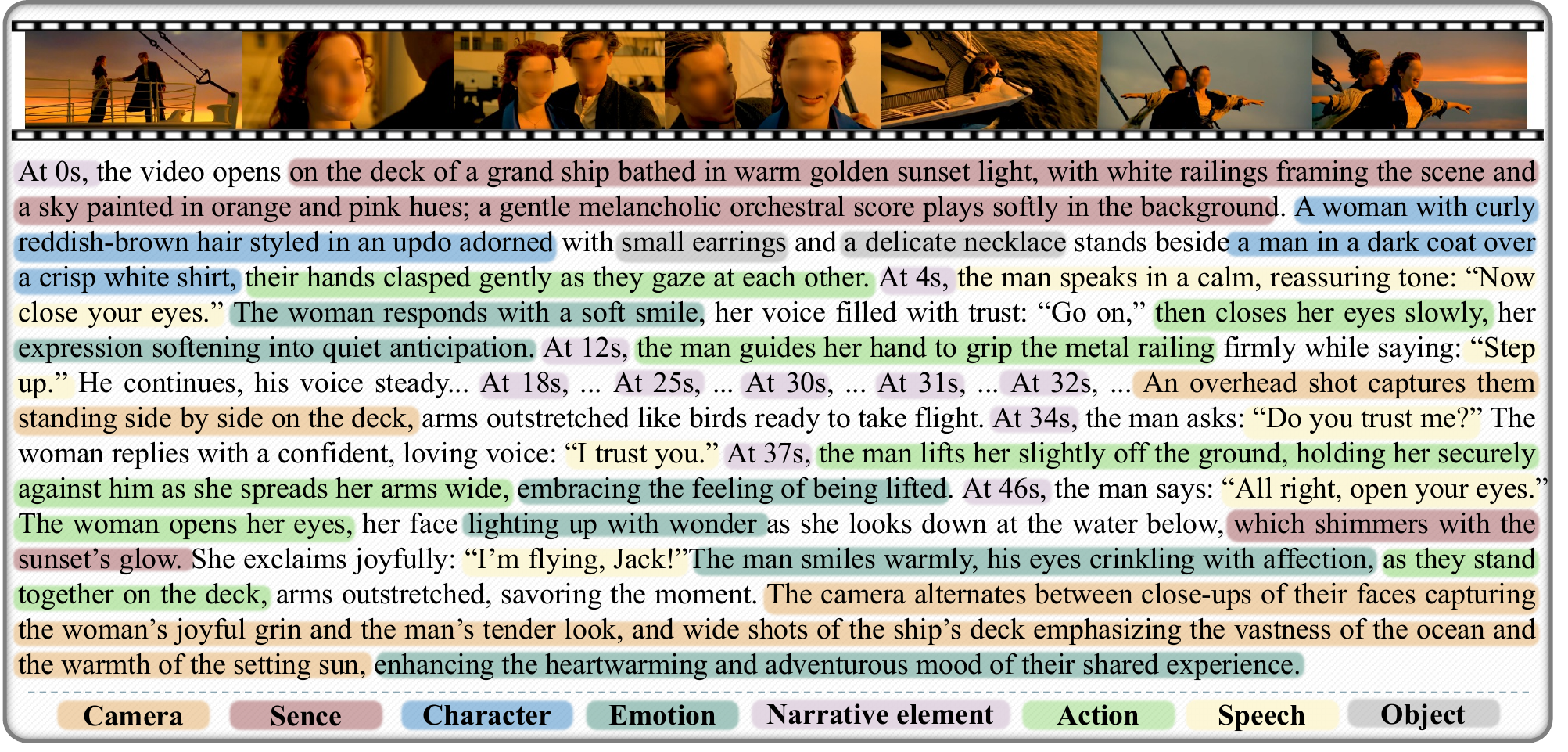}
  \caption{Example of an attribute-structured audiovisual caption generated by \model, with timestamps and grounded speech; color highlights indicate the corresponding attribute groups.}
  \label{fig: demo}
  \vskip -0.1in
\end{figure*}


\begin{table}[t]
  \centering
  \small
  \setlength{\abovecaptionskip}{2pt}
  \setlength{\tabcolsep}{12pt}
  \caption{Stage-wise ablation of the training recipe for captioning.}
  \resizebox{\linewidth}{!}{
  \begin{tabular}{lccccccccc}
  \toprule
      \multirow{2}{*}[-0.1cm]{\textbf{Training Stage}} & \multicolumn{3}{c}{\textbf{video-SALMONN-2 testset}} & \multicolumn{1}{c}{\textbf{DVC Detailed}} \\
      \cmidrule(lr){2-4}
      \cmidrule(lr){5-5}
      ~ & Miss~$\downarrow$ & Hall.~$\downarrow$ & Total~$\downarrow$ &  ~Acc~/~Score~~$\uparrow$ \\
      \midrule
      S1 & 42.1&12.8&54.9& 36.1 / 1.6 \\
      S2 & 24.8&19.9&44.7&  40.4 / 1.8\\
      S3 & 23.4&18.3&41.7&41.7 / 1.9 \\
  \bottomrule
  \end{tabular}
  }
  \addtocounter{footnote}{-1}
  \label{tab:abla training_stages}
  \vskip -0.1in
\end{table}

\subsection{Ablation Study}
All ablation experiments are conducted by randomly sampling 20K training instances and fine-tuning Qwen2.5-Omni-3B under the same optimization settings.

\myPara{Supervision variants for captioning.}
\tabref{tab:abla dataset} analyzes the effect of different supervision data variants on captioning performance.
Using raw annotations yields only small improvements over the baseline, indicating that unstructured ground-truth captions provide weak supervision.
Non-attribute annotation improves descriptive richness, but still leaves many fine-grained details unmodeled, resulting in incomplete captions.
In contrast, multi-attribute supervision achieves the best overall performance, substantially reducing missing content while improving detailed caption quality, demonstrating the benefit of structured, attribute-level supervision.
It is worth noting that encouraging finer-grained descriptions may cause some previously unannotated but relevant details (e.g., on-screen text) to be penalized as hallucinations under the current evaluation protocol.

\myPara{Stage-wise training ablation.}
\tabref{tab:abla training_stages} reports captioning performance after each training stage, with all stages trained on the full dataset.
Training with S1 yields limited improvement, indicating that single-attribute supervision mainly supports basic attribute grounding but is insufficient for comprehensive captioning.
Introducing S2 leads to clear gains, with reduced missing content and improved detailed caption accuracy, highlighting the importance of all-attribute supervision under short temporal contexts.
S3 further improves performance by reducing residual missing and enhancing caption detail, showing that longer temporal contexts better integrate information without increasing hallucinations.

\begin{table}[t]
  \centering
  \small
  \setlength{\abovecaptionskip}{2pt}
  \setlength{\tabcolsep}{12pt}
  \caption{Comparison of attribute-based instruction following for caption generation.}
  \resizebox{\linewidth}{!}{
  \begin{tabular}{lccccccccc}
  \toprule
  \multirow{2}{*}[-0.1cm]{\textbf{Model}}&\multirow{2}{*}[-0.1cm]{\textbf{Size}}&  \multicolumn{4}{c}{\textbf{Number of attributes}} \\
  \cmidrule(lr){3-6}
  &&\textbf{1$\uparrow$} & \textbf{2$\uparrow$} & \textbf{3$\uparrow$} & \textbf{4$\uparrow$} \\
      \midrule
      \textcolor{gray!100}{Gemini-3-Pro} &0&0&0&2.0&3.0   \\
      \textcolor{gray!100}{Gemini-2.5-Pro} &0&0&0&1.5&2.5 \\
      Qwen2.5-Omni & 3B&20.0&8.0&5.0&6.0\\
      Qwen2.5-Omni & 7B&14.0&2.5&6.5&5.5\\
      AVoCaDO & 7B&0&0&0&0\\
      \model~(Ours) & 3B& \textbf{52.3} & \textbf{63.0}& \underline{9.5}& \textbf{7.5} \\
      \model~(Ours) & 7B& \underline{47.0} & \underline{58.0}& \textbf{9.6}& \underline{6.5} \\
  \bottomrule
  \end{tabular}
  }
  \addtocounter{footnote}{-1}
  \label{tab:abla instructions}
  \vskip -0.1in
\end{table}

\subsection{Attribute-level Instruction Following}
To evaluate attribute-level instruction following, we prompt each model with attribute sets containing one to four attributes, and use Gemini-2.5-Pro as an automatic judge to assess whether the generated captions explicitly satisfy the requested attributes.
As shown in~\tabref{tab:abla instructions}, existing captioning models struggle to follow attribute-specified prompts, exhibiting low accuracy even with one or two attributes.
In contrast, \model~consistently achieves higher instruction following accuracy.
Although trained only with single-attribute and all-attribute supervision, \model~generalizes well to unseen attribute combinations as the number of requested attributes increases.

\subsection{Qualitative Analysis}
\figref{fig: demo} shows a representative caption generated by \model.
The caption is explicitly timestamped and integrates multiple complementary aspects of the video, including visual content, camera cues, actions, and grounded speech, into a coherent and temporally aligned narrative.
More qualitative examples are provided in~\secref{sec:visualizations}.

\section{Conclusions}
In this work, we present \dataset, an open-source collection of composable single- and multi-attribute audiovisual instructions, together with \pipeline, a scalable curation pipeline that produces reliable attribute-structured supervision.
By combining multi-source annotation, post-ensemble consistency verification against the source audiovisual content, and attribute-wise evaluation with targeted refinement, \pipeline~improves fine-grained coverage while suppressing error accumulation.
Trained with a progressive three-stage supervised fine-tuning scheme, \model~produces higher-quality audiovisual captions with improved attribute-level instruction following.
Across captioning benchmarks, instruction following, caption-grounded QA, and temporal grounding evaluations, \model~achieves consistent gains over strong open-source models and remains competitive with Gemini-3-Pro.

{\small
\bibliographystyle{ieee_fullname}
\bibliography{main}
}
\appendix
\newpage
\clearpage 

\section*{Limitations}
%

%
First, \dataset~is constructed via multi-source automatic annotation, ensembling, and verification, and residual noise may remain despite Stage~3 attribute-wise checking, particularly for ASR-dependent speech content and fine-grained timestamps.
Second, parts of our evaluation depend on strong proprietary models as judges or downstream solvers (e.g., GPT-4.1/GPT-4o/Seed-1.6/Gemini), which may introduce evaluator bias and makes absolute scores sensitive to the chosen judge.
Finally, \dataset~focuses on short-form videos, and generalization to longer videos or different domains requires further validation.

\section{Implementation Details}
We summarize the stage-wise training configuration of \model~in~\tabref{tab:training_recipe}.
All experiments are conducted based on Qwen2.5-Omni and trained via supervised fine-tuning.
Across all stages, we use AdamW as the optimizer with a learning rate of $2\times10^{-5}$ and a global batch size of 128.
To control computational cost, we enforce a fixed pixel budget per sample, with a reduced total pixel budget in Stage~3.
For optimization efficiency and memory scalability, we adopt DeepSpeed with ZeRO-2 in Stages~1 and~2, and ZeRO-3 offloading in Stage~3.
During Stages~1 and~2, training jointly optimizes the projector modules and the language model, while Stage~3 adopts LoRA for parameter-efficient adaptation.
All models are trained using 32 NVIDIA A800 GPUs with 80\,GB memory each, and inference is conducted on a single NVIDIA A800 GPU.

\begin{table}[ht]
\centering
\footnotesize
\caption{Training-stage configuration of \model.}
\setlength{\tabcolsep}{2pt}
\resizebox{\linewidth}{!}{
\begin{tabular}{lccc}
\toprule
\textbf{Stage} & \textbf{Stage 1} & \textbf{Stage 2} & \textbf{Stage 3} \\
\midrule
\multirow{2}{*}{Purpose}& {Single-Attribute} &{Short-Context} & {Long-Context} \\
& Alignment & Modeling&Modeling \\
\midrule
Video duration & 0 - 3min & 0 - 30s & 30s - 3min \\
Dataset Items & 150k & 136k & 53k \\
Total Pixels & 20M & 20M & 12M \\
Batch size & 128 & 128 & 128 \\
Optimizer & AdamW & AdamW & AdamW \\
Learning rate & 2e-5 & 2e-5 & 2e-5 \\
Warmup Ratio & 5\% & 5\% & 5\% \\
Precision&  Bfloat16 & Bfloat16 & Bfloat16\\
Sequence Length & 16384 & 16384 & 22384 \\
Deepspeed & zero2 & zero2 & zero3\_offload \\
Train Components & Proj., LLM & Proj., LLM & LoRA \\
\bottomrule
\end{tabular}
}
\label{tab:training_recipe}
\end{table}

\section{Annotation Datasets}
\myPara{LLaVA-Video-178K.}
LLaVA-Video-178K~\cite{zhang2024video} is a large-scale open-domain video instruction dataset constructed from YouTube, consisting of untrimmed video clips with durations ranging from 5 seconds to 3 minutes.
The videos span diverse content categories, including daily activities, instructional videos, cooking, sports, TV shows, and egocentric scenarios, and are curated to emphasize temporal diversity and dynamic scenes rather than static content.
From this dataset, we sample approximately 119K videos from the YouTube subset for annotation.

\myPara{FineVideo.}
FineVideo~\cite{Farr2024FineVideo} is a large-scale video dataset consisting of approximately 43K videos with a total duration of about 3.4K hours.
The dataset is curated through automatic filtering to retain videos with dynamic visual content and dense spoken language, using criteria such as visual motion intensity and word-level speech density.
It emphasizes videos with clear action dynamics and temporally rich audiovisual signals, while excluding static scenes and low-information content.
These properties make FineVideo particularly suitable for learning fine-grained temporal alignment between visual events and speech.
From this dataset, we sample approximately 16K short-form videos with durations under three minutes for annotation.

\section{Evaluation Benchmarks}

\myPara{video-SALMONN-2.}
The video-SALMONN-2 test~\cite{tang2025video} contains 483 videos drawn from 14 different domains.
Each video lasts between 30 and 60 seconds, with an average duration of 51 seconds.
Caption quality is evaluated by an automatic judge that compares generated captions with reference events and categorizes errors into missing, incorrect, and hallucinated events.
Incorrect and hallucinated events are treated as hallucination errors, and the final error score is computed by summing the missing rate and the hallucination rate.

\myPara{UGC-VideoCap.}
UGC-VideoCap~\cite{wu2025ugc} consists of 1,000 short-form user-generated videos collected from TikTok.
All videos are shorter than 60 seconds and contain at least one meaningful audio segment lasting no less than five seconds.
Captions are evaluated by a judge model that assigns scores on a five-point scale across three dimensions: visual content, audio content, and descriptive detail.
The scores are normalized and aggregated to produce an overall caption quality score.

\myPara{VDC.}
The VDC~\cite{chai2024auroracap} benchmark contains 1,027 videos and is designed for evaluating fine-grained visual captioning under a structured protocol.
For each video, models generate captions conditioned on five predefined visual aspects, including camera, short summary, background, main object, and detailed description, using aspect-specific prompts.
Performance is measured using both accuracy and judge scores for each aspect.

\myPara{VidCapBench-AE.}
VidCapBench-AE~\cite{chen2025vidcapbench} is an automatically evaluable subset of VidCapBench, designed to assess whether video captions provide sufficient and accurate information for controllable text-to-video generation.
The benchmark is built on a large collection of short video clips drawn from diverse sources, covering a wide range of visual styles, motions, and scene compositions, with most videos being short-form clips suitable for generation-oriented evaluation.
VidCapBench-AE reformulates caption evaluation as aspect-specific question answering, where each video is associated with multiple QA pairs targeting four generation-critical aspects: video aesthetics, video content, video motion, and physical laws.
For each aspect, caption quality is evaluated using accuracy, precision, coverage, and conciseness, enabling fine-grained assessment of both correctness and completeness with respect to generation-relevant semantics.

\myPara{Daily-Omni.}
Daily-Omni~\cite{zhou2025daily} is an audiovisual question answering benchmark consisting of 684 videos depicting diverse everyday scenarios collected from multiple platforms.
The benchmark includes 1,197 multiple-choice question-answer pairs spanning six task categories.
In our evaluation setting, generated captions are provided as the sole input to a fixed QA model, and performance is measured by question answering accuracy.

\myPara{WorldSense.}
WorldSense~\cite{hong2025worldsense} focuses on tightly coupled audiovisual reasoning and contains 1,662 temporally aligned audiovisual clips grouped into eight semantic domains.
The dataset includes 3,172 multiple-choice question-answer pairs covering 26 downstream tasks.
We evaluate caption quality by measuring how effectively the generated captions support accurate question answering when used as input to a judge model.

\myPara{Charades-STA.}
Charades-STA~\cite{gao2017tall} is a temporal grounding benchmark built on the Charades dataset, focusing on localizing natural language queries in indoor videos.
Each sample pairs a textual query with its corresponding temporal segment within a video.
We follow the standard data splits, which include approximately 12.4K training samples and 3.7K validation samples.
In our caption-based evaluation setting, temporal localization is performed using captions only, and performance is reported using mean IoU and recall at multiple IoU thresholds.

\section{Additional Experiments}

\myPara{Training Stage Ablation.}
We further analyze the effect of Stage~1 training supervision by comparing models trained with and without the attribute-wise representation learning stage.
As shown in~\tabref{tab:abla attribute}, introducing Stage~1 consistently reduces missing content and improves overall caption reliability on video-SALMONN-2.
Although the impact on hallucination is limited, the improvement in total error indicates that early attribute-wise supervision helps establish more accurate semantic grounding.
Consistent gains are also observed on DVC Detailed, suggesting that Stage~1 benefits fine-grained visual description quality.

\begin{table}[t]
  \centering
  \small
  \setlength{\tabcolsep}{10pt}
  \caption{Ablation of Stage~1 training supervision on audiovisual captioning performance.}
  \resizebox{\linewidth}{!}{
  \begin{tabular}{lccccccccc}
  \toprule
      \multirow{2}{*}[-0.1cm]{\textbf{Stage}} & \multicolumn{3}{c}{\textbf{video-SALMONN-2 testset}} & \multicolumn{1}{c}{\textbf{DVC Detailed}} \\
      \cmidrule(lr){2-4}
      \cmidrule(lr){5-5}
      ~ & Miss~$\downarrow$ & Hall.~$\downarrow$ & Total~$\downarrow$ &  ~Acc~/~Score~~$\uparrow$ \\
      \midrule
      w/o S1 & {27.2} & {19.3}& {46.5}& {40.0} / {1.7}\\
      w/ S1  & \textbf{25.5} & \textbf{18.9}& \textbf{43.4}& \textbf{40.2} / \textbf{1.7} \\
  \bottomrule
  \end{tabular}
  }
  \addtocounter{footnote}{-1}
  \label{tab:abla attribute}
\end{table}

\section{Attribute Coverage Analysis}
\begin{figure*}
    \centering
    \includegraphics[width=1\linewidth]{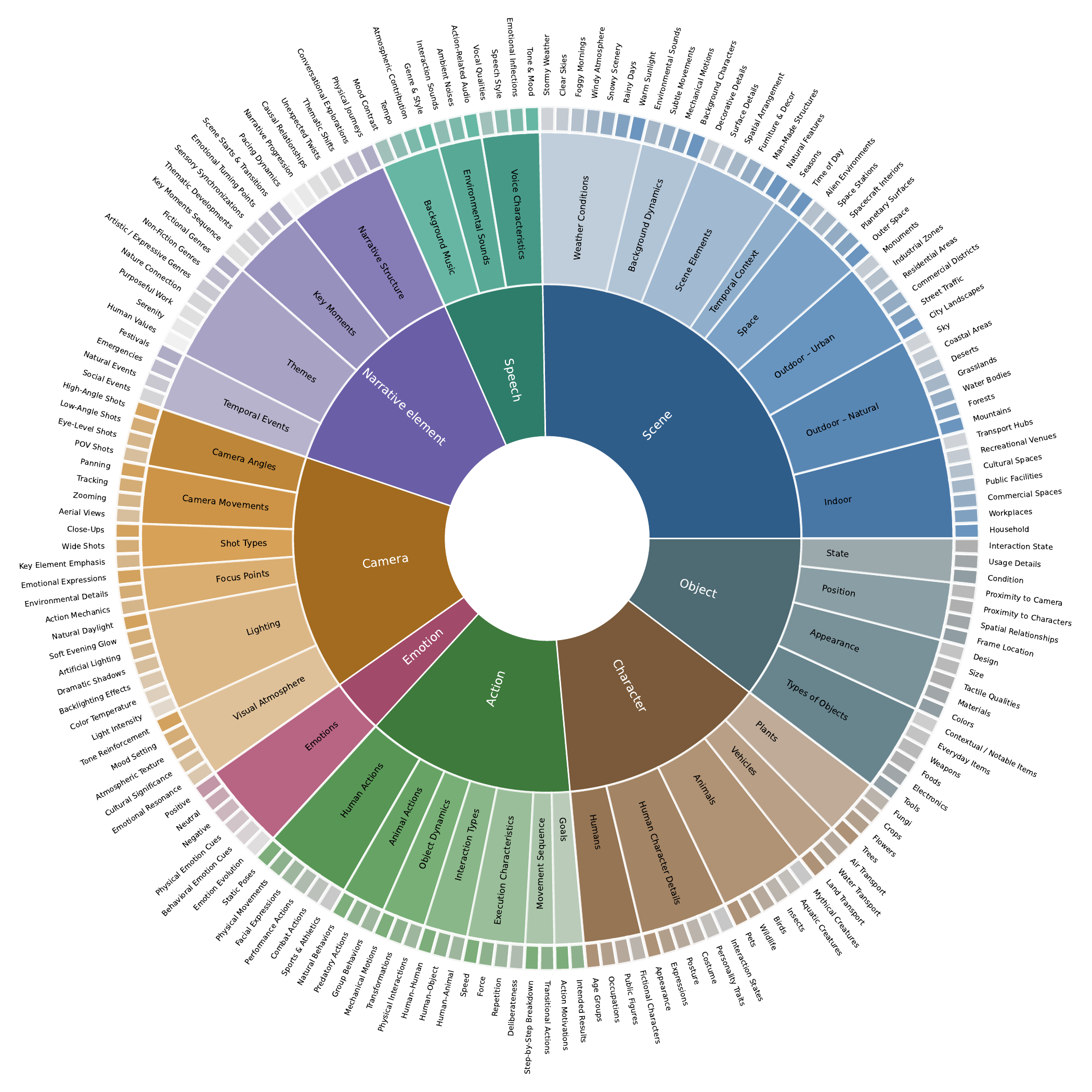}
    \caption{Attribute taxonomy and dataset-wide coverage statistics.
    The sunburst visualizes the hierarchical attribute taxonomy used in our dataset, where each leaf node corresponds to a fine-grained attribute.
    }
    \label{fig:sunburst}
\end{figure*}

\figref{fig:sunburst} presents the hierarchical organization of fine-grained semantic categories used in our analysis.
The hierarchy consists of three levels, progressing from coarse to fine granularity.
The first two levels are manually defined and serve as the annotation schema during dataset construction, specifying high-level and mid-level semantic dimensions such as scene, action, object, and speech.
Based on the completed annotations, we further derive the third, finest-grained level by analyzing the attribute-structured captions.
Based on the completed annotations, the third-level categories are automatically derived via semantic grouping.
For each second-level category, Seed-1.6 groups the corresponding caption spans into fine-grained categories.
GPT-5.2 is then used for post-processing to merge redundant categories and normalize names.

\section{Visualizations}
\label{sec:visualizations}

\begin{figure*}
    \centering
    \includegraphics[width=1\linewidth]{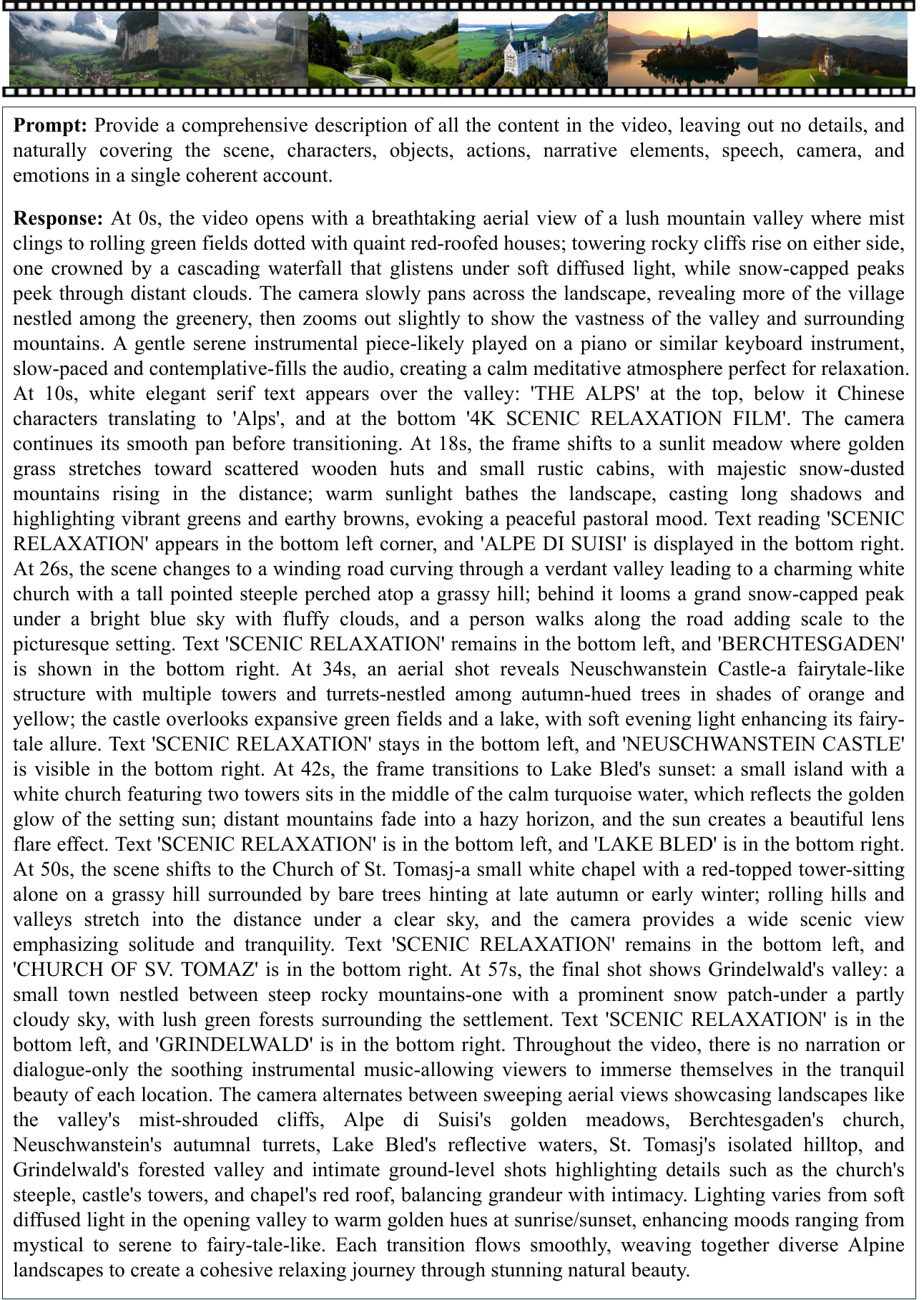}
    \caption{Visualization of captions generated by the \model.}
    \label{fig:demo1}
\end{figure*}

\begin{figure*}
    \centering
    \includegraphics[width=1\linewidth]{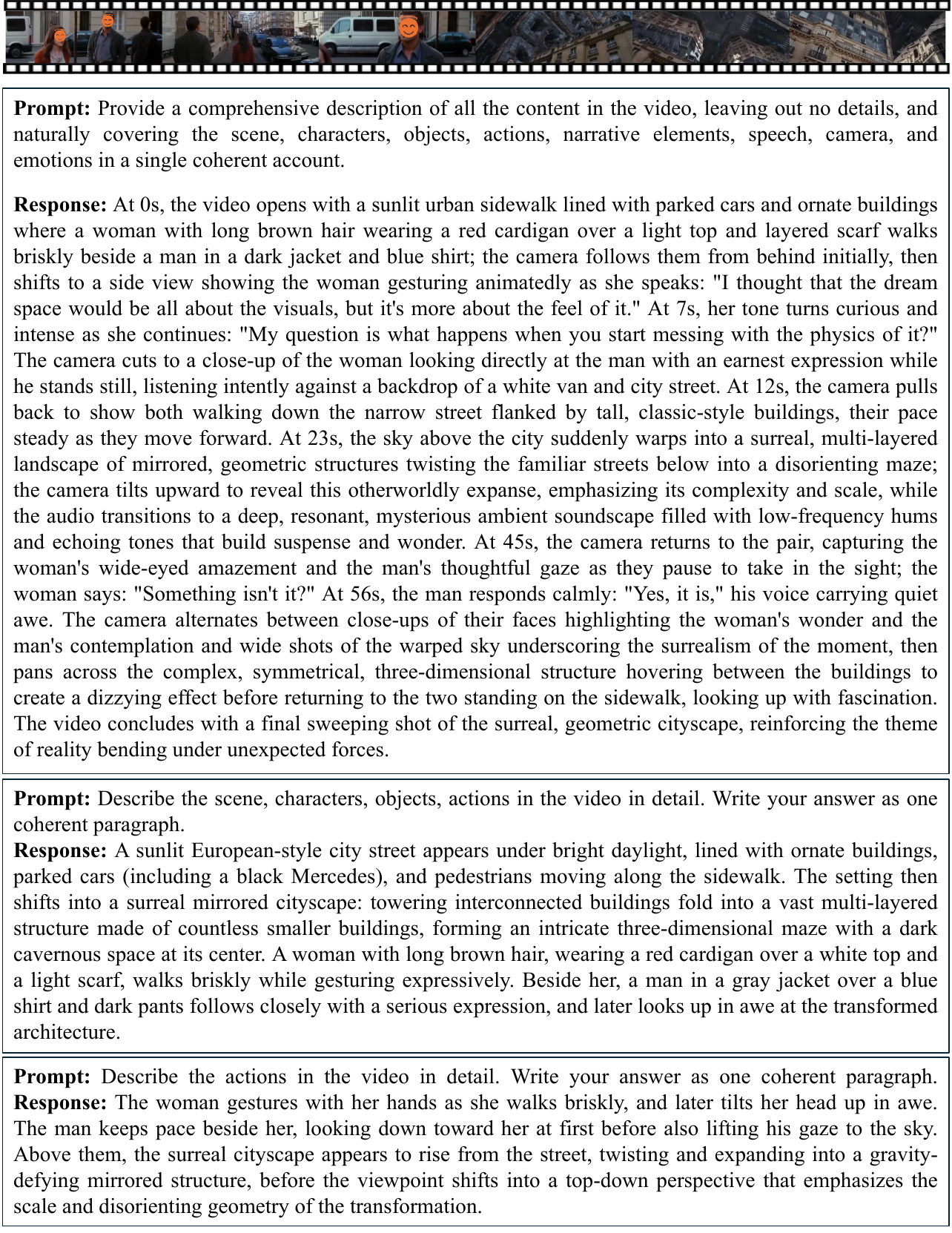}
    \caption{Visualization of captions generated by the \model.}
    \label{fig:demo2}
\end{figure*}

\begin{figure*}
    \centering
    \includegraphics[width=1\linewidth]{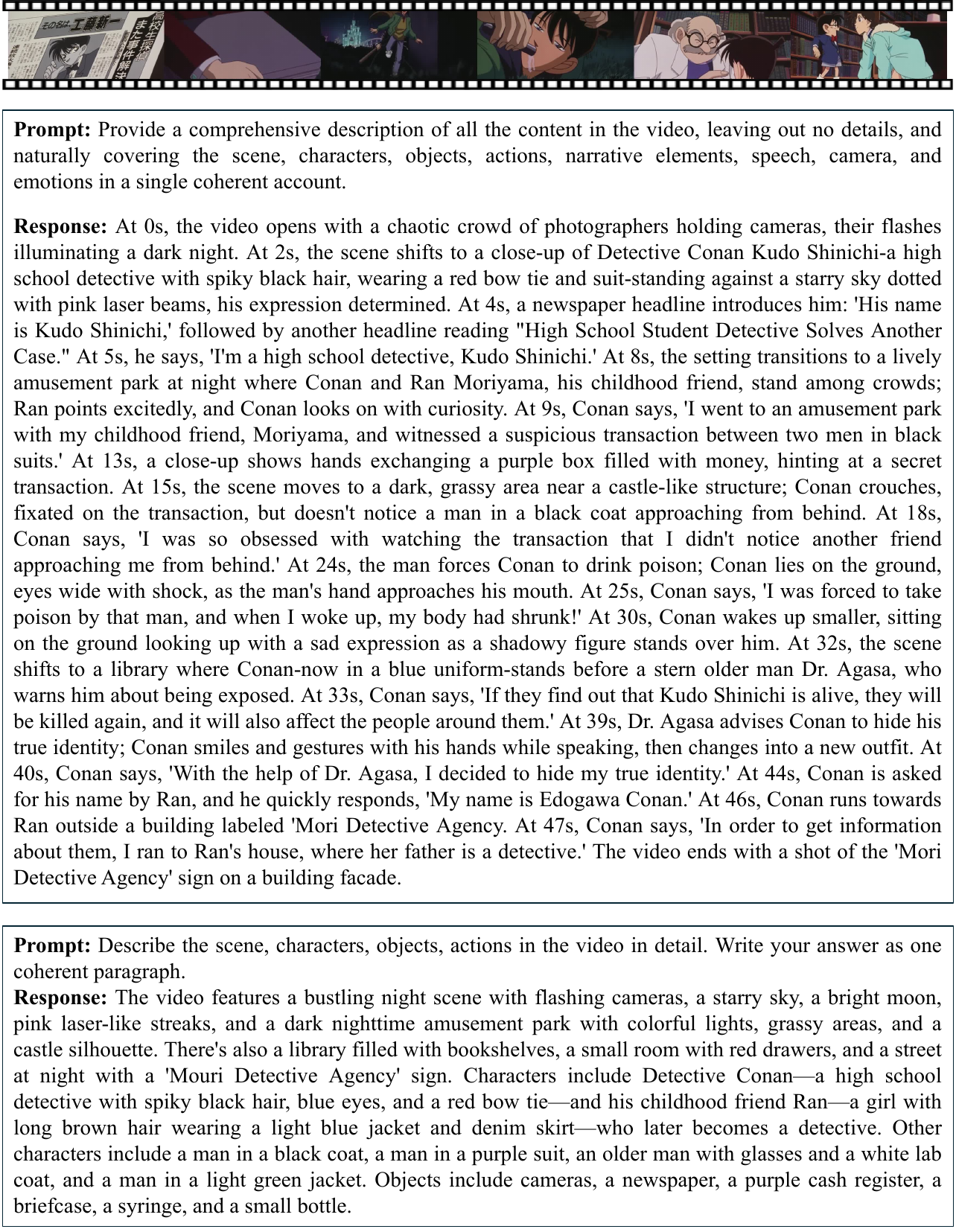}
    \caption{Visualization of captions generated by the \model.}
    \label{fig:demo3}
\end{figure*}

\begin{figure*}
    \centering
    \includegraphics[width=1\linewidth]{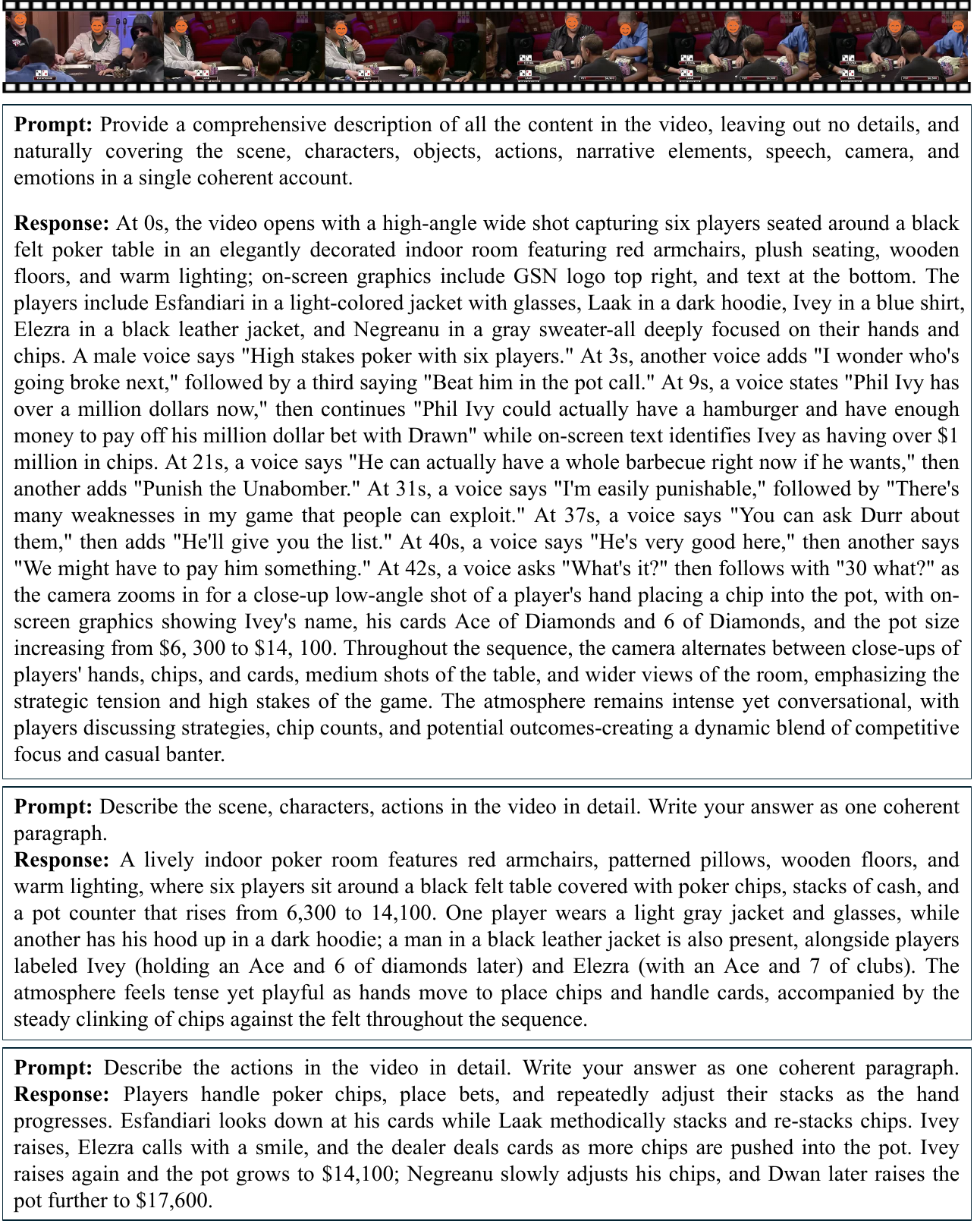}
    \caption{Visualization of captions generated by the \model.}
    \label{fig:demo4}
\end{figure*}

\begin{figure*}
    \centering
    \includegraphics[width=1\linewidth]{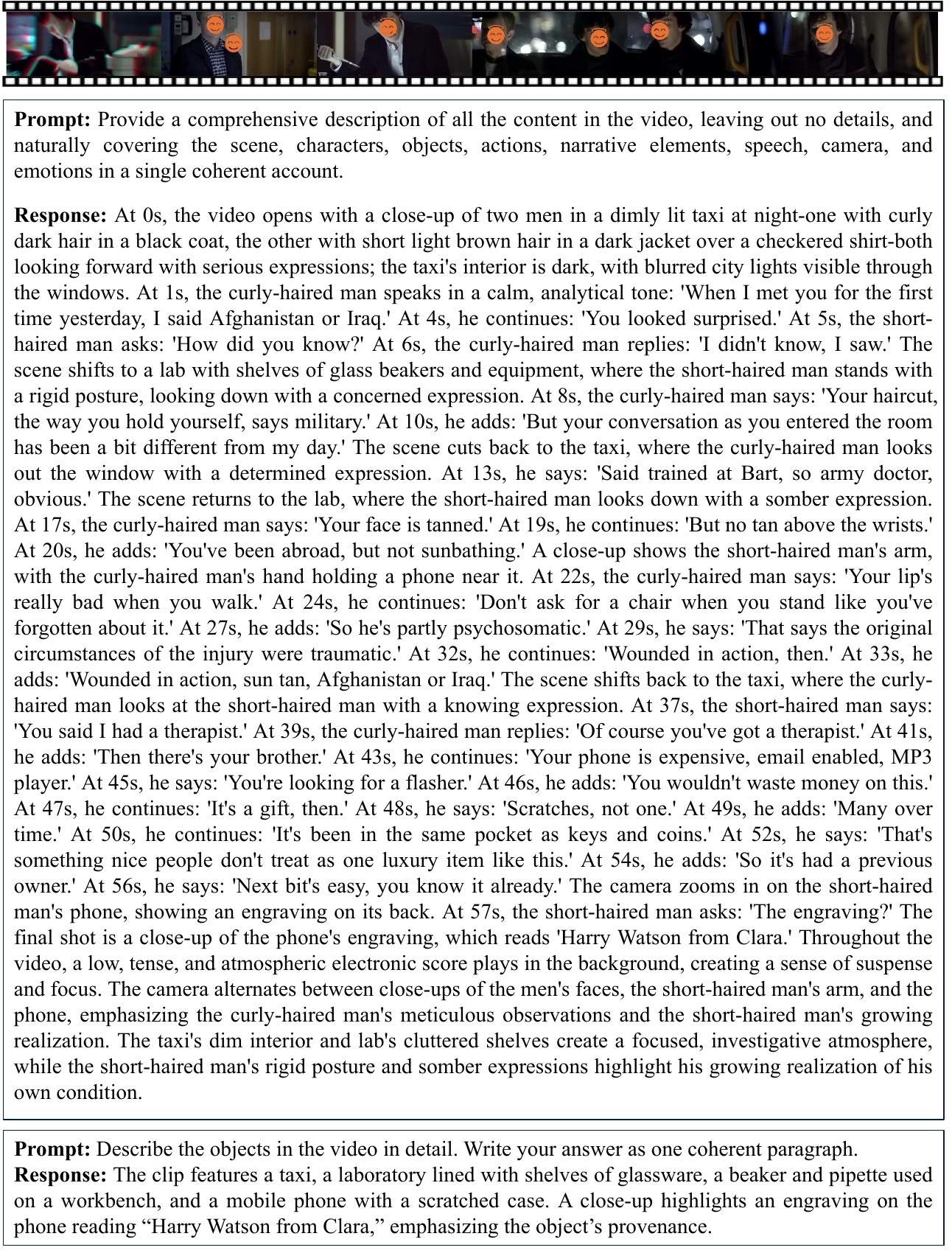}
    \caption{Visualization of captions generated by the \model.}
    \label{fig:demo5}
\end{figure*}

\begin{figure*}
    \centering
    \includegraphics[width=1\linewidth]{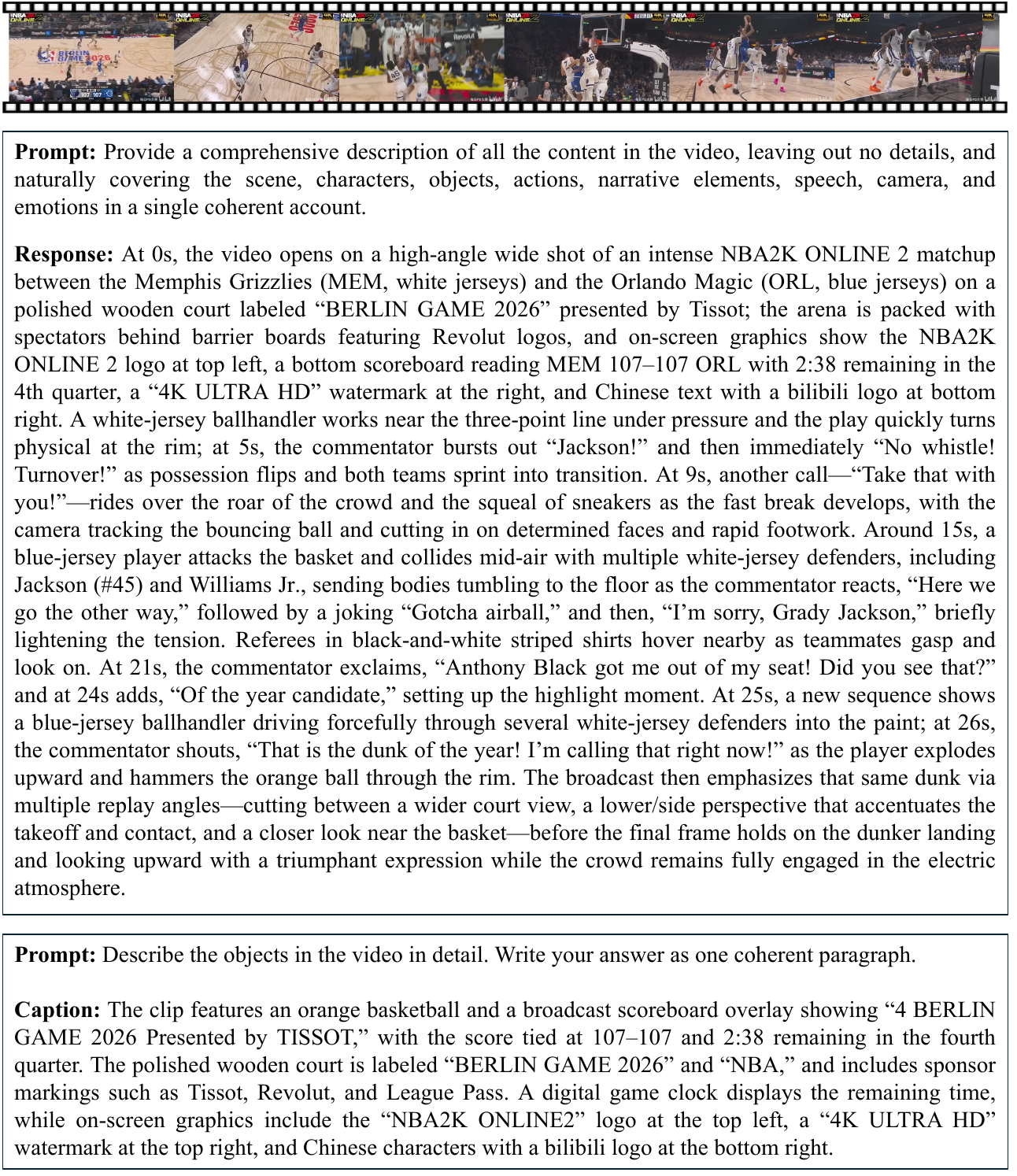}
    \caption{Visualization of captions generated by the \model.}
    \label{fig:demo6}
\end{figure*}

\begin{figure*}
    \centering
    \includegraphics[width=1\linewidth]{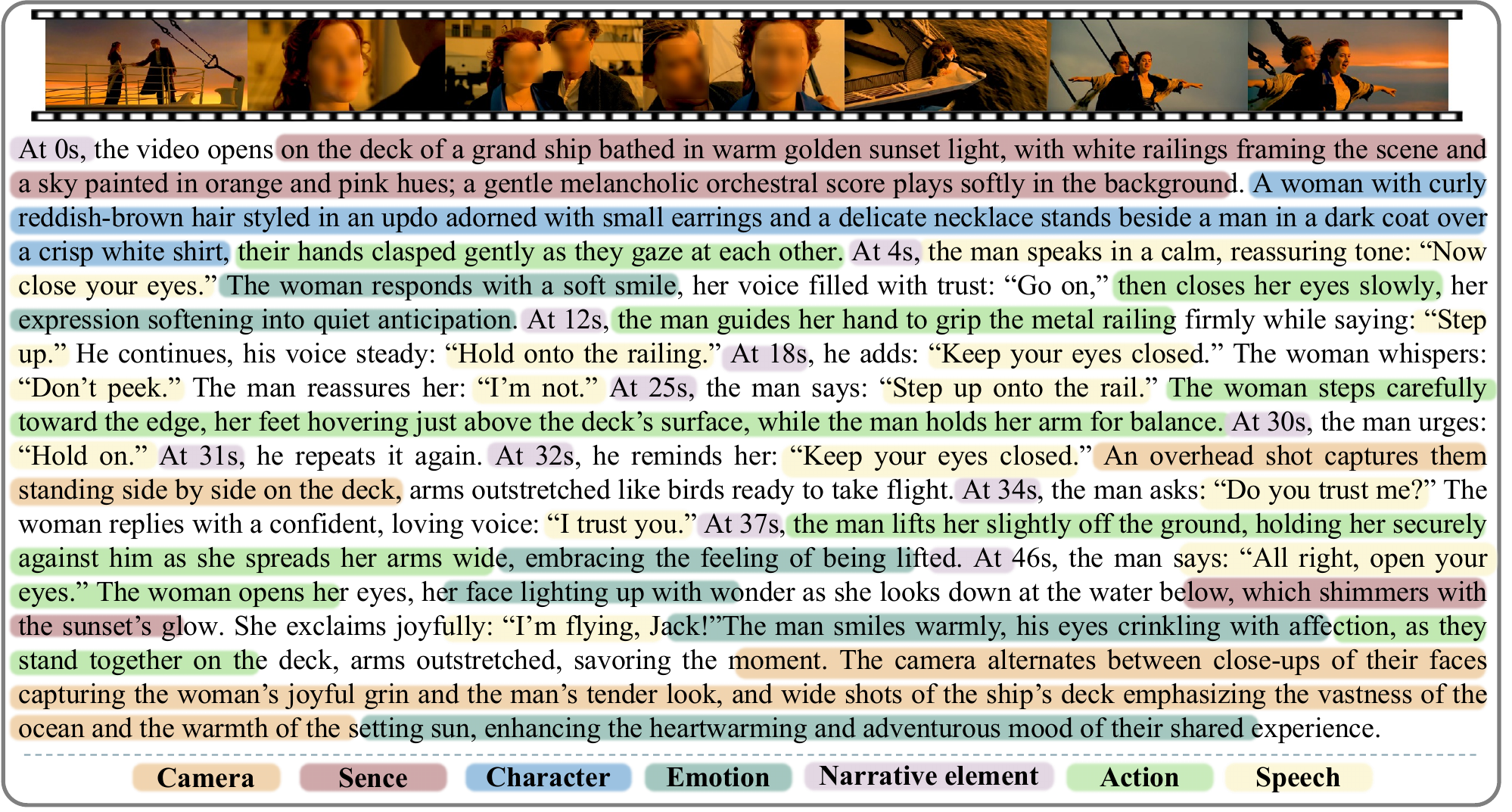}
    \caption{Visualization of captions generated by the \model.}
    \label{fig:demo7}
\end{figure*}

\figref{fig:demo1}-\figref{fig:demo7} visualize captions generated by the ASID-Captioner across diverse real-world video scenarios, including natural landscapes, urban scenes, narrative content, human interactions, and sports footage.
The results demonstrate that the model produces fine-grained, temporally coherent, and well-grounded captions under different instruction settings, highlighting its robustness and generality in realistic settings.

\section{Prompts}
We summarize the prompts used throughout the data pipeline and evaluation.
For data annotation, we first generate detailed timestamped captions from subtitle-annotated clips (\figref{fig:subtitle_caption_prompt}) and ensemble multi-source captions into a single coherent narrative (\figref{fig:merge_prompt}).
For quality control, we verify and complete captions by identifying important visual and speech deficiencies with structured JSON outputs (\figref{fig:verify_complete_prompt}), followed by attribute-wise auditing to report only verifiable errors, missing items, and expressiveness issues (\figref{fig:caption_audit_prompt}), and audit-guided refinement to minimally fix reported issues while preserving the original timestamp structure (\figref{fig:caption_refine_prompt}).
For evaluation, we use caption-grounded multiple-choice QA (\figref{fig:qa_judge}) and caption-based temporal grounding (\figref{fig:ground_judge}) to assess factual coverage and temporal localization from textual captions.





\begin{figure*}
  \begin{tcolorbox}[colback=white!95!gray, colframe=gray!50!black, rounded corners, title={Prompt template for generating detailed timestamped captions from subtitle-annotated clips.}]
  You are a video captioning assistant.\\
  
  You will be given a video clip with (i) spoken dialogue subtitles tagged with speaker names and time ranges, and (ii) separate timestamp cues.\\
  
  Task: Write ONE comprehensive paragraph that seamlessly integrates \textbf{Scene, Characters, Objects, Actions, Narrative (with multiple in-sentence timestamps in chronological order), Speech (as direct quotes), Camera, and Emotions}.\\
  
  Rules:
  (1) Use ONLY information supported by the clip/subtitles.\\
  (2) Include key moments with explicit timestamps (in seconds) embedded directly in sentences; reflect scene transitions, turning points, and pacing.\\
  (3) Do NOT mention subtitles, on-screen text, or any flashing/overlay elements.\\
  (4) Do NOT use speaker IDs; indicate speakers naturally with dialogue tags (e.g., ``she says'').\\
  (5) No lists, brackets/parentheses, or bullet-like formatting—output a single flowing paragraph with rich detail.\\
  
  Inputs: \{subtitle-annotated clip\}\\
  Output: \{one-paragraph timestamped detailed caption\}
  \end{tcolorbox}
  \caption{Prompt template for generating detailed timestamped captions from subtitle-annotated clips.}
  \label{fig:subtitle_caption_prompt}
\end{figure*}

\begin{figure*}
  \begin{tcolorbox}[colback=white!95!gray, colframe=gray!50!black, rounded corners, title={Prompt template for ensembling multi-source video captions.}]
  You are a video narrative integrator. You will be given three textual descriptions of the same video from different sources.\\
  
  Task: Merge them into ONE seamless English paragraph that preserves all specific details, maintains chronological order, and synchronizes visual events with corresponding audio/dialogue.\\
  
  Timestamp rule: Include only key moments using \textbf{single integer-second anchors} in the form ``At Xs'' (no ranges, decimals, parentheses, or duplicate timestamps), spaced naturally across the narrative.\\
  
  Constraints: Do not mention source names or generation process; do not invent information; keep Whisper dialogue verbatim; output a single paragraph (no lists or brackets).\\
  
  Inputs:
  Description 1: \{seed\}\\
  Description 2: \{avocado\}\\
  Description 3: \{whisper\}\\
  Output: \{one-paragraph merged narrative\}
  \end{tcolorbox}
  \caption{Prompt template for ensembling multi-source video captions.}
  \label{fig:merge_prompt}
\end{figure*}

\begin{figure*}
  \begin{tcolorbox}[colback=white!95!gray, colframe=gray!50!black, rounded corners, title={Prompt template for caption verification and one-pass completion.}]
  You are a video caption verification and completion assistant.\\
  
  Inputs: (1) CURRENT\_CAPTION, and (2) SOURCES (Whisper ASR + multi-source captions).\\
  
  Task: (i) identify \emph{important} missing content, separating purely visual deficiencies from speech deficiencies (speech only from Whisper), and (ii) produce a one-time completed caption by supplementing only supported missing details.\\
  
  Constraints: no fabrication; ignore trivial details; keep timestamps as integer seconds and follow the timestamp style of CURRENT\_CAPTION; output \textbf{valid JSON with exactly three keys}:\\
  \{\texttt{non\_speech\_deficiency\_specific}, \texttt{speech\_deficiency\_specific}, \texttt{caption}\}.\\
  
  CURRENT\_CAPTION: \{current\_caption\}\\

  SOURCES: \{whisper, seed, avocado\}
  \end{tcolorbox}
  \caption{Prompt template for caption verification and one-pass completion.}
  \label{fig:verify_complete_prompt}
\end{figure*}

\begin{figure*}
  \begin{tcolorbox}[colback=white!95!gray, colframe=gray!50!black, rounded corners, title={Prompt template for attribute-wise caption auditing.}]
  You are a caption auditor. You are given the video, optional ASR (truth source for speech meaning only), and a caption to audit.\\
  
  Task: Report \textbf{shortcomings only} (errors, missing, expressiveness), and \textbf{audit ONLY the items explicitly listed in each dimension's VERIFY list}; do not evaluate anything outside the VERIFY scope.\\
  
  Evidence rule: output an error only when a contradiction is clear and verifiable. Missing is reported only for clearly present, important items under VERIFY.\\
  
  Output: Return \textbf{strict JSON only} following the provided schema (no markdown, no explanations). Use:
  errors as \texttt{\{"snippet": "...", "why": "..."\}}, missing as \texttt{\{"what": "..."\, "why": "..."\}}, and expressiveness as a list of short strings.\\
  
  Inputs: \{video\}, \{asr (optional)\}, \{caption\}
  \end{tcolorbox}
  \caption{Prompt template for attribute-wise caption auditing.}
  \label{fig:caption_audit_prompt}
\end{figure*}

\begin{figure*}
  \begin{tcolorbox}[colback=white!95!gray, colframe=gray!50!black, rounded corners, title={Prompt template for audit-guided caption refinement.}]
  You are a caption refiner. You are given a caption and its audit report.\\
  
  Task: Produce an improved caption by (i) fixing factual errors, (ii) filling important missing items, and (iii) doing minimal language-only polishing for clarity.\\
  
  Constraints: Use the audit report as the only edit specification; do not add new content beyond it. Keep the original timestamp blocks and their order unchanged. Timestamps must follow ``At Xs,'' (X is integer or one-decimal), with no ranges.\\
  
  Output: Return \textbf{strict JSON only} with keys \texttt{improved\_caption} and \texttt{enhance\_summary} (fixed\_errors, filled\_missing, dropped\_forbidden, timestamp\_adjustments).\\
  
  Inputs: \{caption\}, \{audit\_report\}
  \end{tcolorbox}
  \caption{Prompt template for audit-guided caption refinement.}
  \label{fig:caption_refine_prompt}
\end{figure*}

\begin{figure*}
  \begin{tcolorbox}[colback=white!95!gray, colframe=gray!50!black, rounded corners, title={Prompt template for multiple-choice question answering based on textual video captions.}]
  You are a precise QA assistant. Your task is to answer multiple-choice questions based ONLY on the video caption provided.\\
  
  Do not use any outside knowledge or assumptions—your answer must strictly reflect information from the caption. 
  Always output only the capital letter corresponding to your choice (e.g., A, B, C, D). 
  If the caption does not provide enough information to answer the question, output "N/A" instead.\\
  
  Here is the video caption: \{video caption\}\\
  
  Question: \{question\}
  
  Choices: \{choices\}
  \end{tcolorbox}
  \caption{Prompt template for multiple-choice question answering based on textual video captions.}
  \label{fig:qa_judge}
  \end{figure*}

\begin{figure*}
  \begin{tcolorbox}[colback=white!95!gray, colframe=gray!50!black, rounded corners, title={Prompt template for temporal grounding based on textual video captions.}]
  You are a temporal grounding assistant.\\

  You will be given:
  (1) a long video caption with multiple timestamp anchors like "At 0s, ... At 5s, ...",
  (2) an event description (a sentence).\\

  Goal:
  Infer the most likely continuous time interval (start and end in seconds) when the event happens, using ONLY the caption.\\

  Rules:
  1) Always try to output a time interval, even if the evidence is partial. Use best-effort inference.
  2) Prefer intervals aligned to existing anchors. If the event is mentioned near "At Ts", choose a range that covers that anchor and the most plausible neighboring anchors.
  3) If the event is implied by related actions/objects (synonyms/paraphrases), still infer the interval by matching the closest described segment.
  4) Output "N/A" ONLY if the caption provides absolutely no usable clue to localize the event (no matching action/object/context anywhere).\\
  
  Here is the video caption: \{video caption\}\\
  
  Question: \{question\}
  
  Choices: \{choices\}
  \end{tcolorbox}
  \caption{Prompt template for temporal grounding based on textual video captions.}
  \label{fig:ground_judge}
  \end{figure*}



\end{document}